\documentclass[pdflatex,sn-mathphys]{sn-jnl}

\theoremstyle{thmstyleone}%
\usepackage[justification=centering]{caption}
\usepackage{commath}
\usepackage{float}
\theoremstyle{thmstyletwo}%

\theoremstyle{thmstylethree}%

\begin{document}

\title[Enhanced Opposition Differential Evolution Algorithm for Multimodal Optimization]{Enhanced Opposition Differential Evolution Algorithm for Multimodal Optimization}


\author[*]{\fnm{Shatendra} \sur{Singh}}

\author[]{\fnm{Aruna} \sur{Tiwari}}

\affil[1]{singh.shatendra@gmail.com}
\affil[]{artiwari@iiti.ac.in}


\affil{\orgdiv{Computer Science Engineering}, \orgname{IIT Indore}, \orgaddress{ \city{Indore},  \state{Madhya Pradesh}, \country{India}}}



\abstract{Most of the real-world problems are multimodal in nature that consists of multiple optimum values. Multimodal optimization is defined as the process of finding multiple global and local optima (as opposed to a single solution) of a function. It enables a user to switch between different solutions as per the need while still maintaining the optimal system performance. Classical gradient-based methods fail for optimization problems in which the objective functions are either discontinuous or non-differentiable. Evolutionary Algorithms (EAs) are able to find multiple solutions within a population in a single algorithmic run as compared to classical optimization techniques that need multiple restarts and multiple runs to find different solutions. Hence, several EAs have been proposed to solve such kinds of problems. However, Differential Evolution (DE) algorithm is a population-based heuristic method that can solve such optimization problems, and it is simple to implement. The potential challenge in Multi-Modal Optimization Problems (MMOPs) is to search the function space efficiently to locate most of the peaks accurately. The optimization problem could be to minimize or maximize a given objective function and we aim to solve the maximization problems on multimodal functions in this study. Hence, we have proposed an algorithm known as Enhanced Opposition Differential Evolution (EODE) algorithm to solve the MMOPs. The proposed algorithm has been tested on IEEE Congress on Evolutionary Computation (CEC) 2013 benchmark functions, and it achieves competitive results compared to the existing state-of-the-art approaches.}




\maketitle

\section{Introduction}\label{sec1}

Evolutionary algorithms (EAs) are population-based approaches that have been widely used for solving optimization problems having a single global optimum. It is better to find several optima so that if an optimal solution cannot be used due to resource constraints, other equally good solutions can be used enabling the system performance stay at peak.  Also, the knowledge of multiple optimal solutions in the search space enables a user to get useful insights into the properties of optimal solutions of the function. Many real-world problems consist of multiple optima, such as holographic design\cite{app1}, electro-magnetic design\cite{app2}, protein structure prediction \cite{app3}, and data mining \cite{app4}. Therefore, it is essential to find as many global optimizers as possible as it gives users the flexibility to choose alternate solutions if one solution cannot be chosen due to limited resource constraints. However, it is more challenging to find multiple global optima than to find a single global optimum. Many researchers have used EAs\cite{EAs} and
swarm intelligence algorithms\cite{Swarm} to solve Multi-Modal Optimization Problems (MMOPs), such
as the Genetic Algorithm (GA) \cite{GA}, Ant Colony Optimization (ACO) \cite{ACO}, Estimation of Distribution Algorithm (EDA)\cite{EDA}, Particle Swarm Optimization (PSO)\cite{PSO}, and Differential Evolution (DE)\cite{DE}. However, they tend to lose the effective balance between exploration and exploitation of the search space. Exploration here means increasing the diversity of the population in the hope to find better individuals while the term Exploitation here means refining the existing candidate solutions to reach the best one in a local region(sub-population). While solving the exploration vs exploitation dilemma, two major problems arise. First, if diversity is increased too much then it may lead to unnecessary fitness computations as there may be no optima in those areas of search space where the search is being performed. Second, if exploitation is performed in the undesired region of the search space, then it may get trapped in local optima.
Therefore, a different mechanism, based on classical EAs, is required to locate multiple optima simultaneously. Niching \cite{niching} has been widely used in the literature to help
an EA maintain population diversity in multimodal optimization. Some well-known niching techniques include
crowding \cite{niching1}, clearing\cite{niching2}, fitness sharing \cite{niching3}, and
speciation \cite{niching4}. Niching divides the population into several sub-populations and each subpopulation is responsible for finding the optima within that sub-population. Each sub-population can be considered as a species. Since the fitness landscape may be uneven and complex, the niches could be of different shapes and sizes. Therefore, it becomes a great challenge to form multiple sub-populations out of the global population in such a way that a maximum number of peaks are located by the sub-populations. Classical niching methods are sensitive to the parameters they use; for example, crowding is sensitive to the crowding size, and speciation is sensitive to the niching radius. Hence, parameter-free or parameter-insensitive techniques have been developed to improve niching, such as history-based topological speciation \cite{topological}, and clustering \cite{clustering}. Other adaptive learning strategies have been developed to enhance the diversity for EAs, such as GA\cite{ada1}, DE\cite{ada2}, and PSO\cite{ada3}.

\section{Preliminaries}

\subsection{Differential Evolution}
DE  is a  stochastic method, and it was first introduced in 1997 by Storn and Price\cite{DE}. It has been used for solving a wide variety of problems, including multi-objective, constrained, dynamic, and multimodal optimization problems. DE consists of several operations such as initialization, mutation, crossover, and selection, similar to operations in traditional evolutionary algorithms. The population is initialized randomly within the bounds of the dimensions of the problem. Mutation is the process of changing the genetic sequence of individuals. DE uses vector difference-based mutation operation, which perturbs the current generation population members with a scaled difference of randomly selected and distinct population members.
\begin{equation}
\vec{V}_{i, G}=\vec{X}_{r_{1}, G}+F * (\vec{X}_{r_{2}, G}-\vec{X}_{r_{3}^{j}, G})
\end{equation}

$\vec{V}_{i, G}$ is the offspring generated by adding parent $\vec{X}_{r_{1}, G}$ by scaled vector difference of parents $\vec{X}_{r_{2}, G}$ and $\vec{X}_{r_{3}^{j}, G}$.
After mutation, DE generates a trial vector $\vec{u}_{i, G}= u_{1, i, G}, u 2, i, G, \ldots, u_{D, i, G}$ by crossover operation to enhance the potential diversity of the population. In the basic version, a binomial crossover operation is employed by DE as
\begin{equation}
u_{j, i, G}=\left\{\begin{array}{c}
v_{j, i, G}, \text { if }\left(\operatorname{rand}_{i, j}[0,1] \leq \mathrm{CR}\right) \text { or }\left(j=j_{\text {rand }}\right) \\
x_{j, i, G,}\indent \text {otherwise\hspace{4cm}}
\end{array}\right.
\end{equation}
where i=1,2, $\ldots$, NP, j=1,2, $\ldots$, D, and the crossover rate (CR) is a user-specified constant between 0 and 1 to control the fraction of parameter values copied from the donor vector. $j_{\text {rand }}$ is an integer randomly chosen from 1 to D, which ensures that $\vec{u}_{i, G}$ gets at least one component from $\vec{v}_{i, G}$.
Selection compares the target vector $\vec{x}_{i, G}$ with the trial vector $\vec{u}_{i, G}$ in terms of the fitness values to determine which vector may survive to the next generation
\begin{equation}
\vec{x}_{i, G+1}=\left\{\begin{array}{ll}
\vec{u}_{i, G}, & \text { if } f\left(\vec{u}_{i, G}\right) \leq f\left(\vec{x}_{i, G}\right) \\
\vec{x}_{i, G}, & \text { otherwise }
\end{array}\right.
\end{equation}
\subsection{ODE Framework}
ODE is a combination of the concept of opposition based learning and the DE algorithm. Let $k \in [ a,b ] $ then opposite number (o) is defined as : $o=a+b-k$\cite{b1}. ODE framework mainly consists of two sections, i.e. opposition-based population initialization and opposition-based generation jumping.
\subsubsection{Opposition-Based Population Initialization}
The randomly generated initial population is used to produce an opposition-based population.
The opposite population is calculated as : $O P_{i, j}=a_{j}+b_{j}-P_{i, j}, i=1,2, \ldots, N_{p} ; j=1,2, \ldots, D,$ where $P_{i,j},$ and
$OP_{i, j}$ denote the $j^{th}$ variable of the $i^{th}$ vector of the population and opposite population, respectively \cite{b1}.
\subsubsection{ Opposition Based Generation Jumping}
This concept helps in the generation of new individuals that are likely to be more fitter than their corresponding parents. Jumping Rate($J_{r}$) is used to control the degree of opposition of the offspring candidates. $J_{r}$ also indirectly controls the diversity of the population. The higher the value of the jumping rate, the more the search space shrinks. The value of $J_{r}$ is chosen to be 0.3 based on \cite{b1}. The search space is reduced in each generation and the bounds of the dimensions are updated according to the generation. These bounds of each generation are used to compute the opposite population using 
\begin{equation}
OP_{i, j}=\mathrm{MIN}_{j}^{p}+\mathrm{MAX}_{j}^{p}-P_{i, j}
\end{equation}
Here $\mathrm{MIN}_{j}^{p},\mathrm{MAX}_{j}^{p}$ represents the minimum and maximum values of the variable in $j^{th}$ dimension in the current generation \cite{b1}.

\subsection{NBC-Minsize}
Nearest Better Clustering (NBC)\cite{NBC} is a tree-based technique used in MMOPs to divide the population into several species and each species tries to find an optimal solution. NBC-Minsize\cite{NBC-Minsize} is an enhanced version of NBC. The parameter $\varphi$, in NBC, is responsible for controlling the number of species in the population. Smaller values of $\varphi$ result in formation of increasingly multiple species whereas higher values result in fewer species. It is important to note that species with few members (e.g. 1 or 2) find it difficult to evolve with the mutation operators of DE.
Therefore, the parameter $minsize$ is used to limit the minimal number of individuals in the species. It is important to note that too small value of $minsize$ makes the process ineffective whereas too large value would result in merging two species as one. To be able to locate multiple peaks, it is wise to allocate small values in the early stages and relatively higher values in the later stages of the evolution. In the early stages, the focus of evolution must be to explore the function space well, and hence multiple species formation is encouraged by using lower values of $minsize$. In the later stages, the focus should be on exploiting the search space and hence a larger $minsize$ value is preferred. Thus, the species converge to the global optima more quickly. Algorithm 1 refers to the NBC-Minsize.

\begin{algorithm}
  \caption{NBC-Minsize}\label{}
  \begin{algorithmic}[1]
      \State Set minsize by equation (5), (6);
      \State Construct the spanning tree T ;
      \State Calculate the mean distance $\mu_{dist}$ ;
      \State Calculate the follow vector;
      \State Sort the edges in T from the longest to the shortest;
      \For {each $e \in T$}
      \State\text{\textbf{if} $dist(e)$ $\textgreater$ $\varphi * \mu_{dist}$ \textbf{then}} 
            \State\hspace{0.5cm} Set $e_{f}$ to the follower individual of e;
            \State\hspace{0.5cm} Set $e_{r}$ to the root of the subtree containing $e_{f}$ ;
            \State\hspace{0.5cm}\text{\textbf{ if} $follow(e_{f})\geq minsize$ \textbf{and}}
            \State\hspace{0.5cm}\text{$follow(e_{r}) - follow(e_{f}$) $\geq$ $minsize$ \textbf{and}}
            \State\hspace{0.5cm}\text{ f($\frac {e_{f}+e_{r}}{2}$)\textless $f(e_{r}$) \textbf{and} f($\frac {e_{f}+e_{r}}{2}$)\textless $f(e_{f}$) \textbf{then}}
            \State\hspace{0.8cm} Cut off $e$;
            \State\hspace{0.8cm} Set $e_{l}$ to the leader individual of e;
            \State\hspace{1cm}{\textbf{for} each $x$ on the path from $e_{l}$ to $e_{r}$ \textbf{do}}
            \State\hspace{1.2cm} 
            $follow(x)=follow(x)-follow(e_{f})$
            \State\hspace{1cm}\textbf{end for}
            \State\hspace{0.8cm}\textbf{end if}
            \State\hspace{0.11cm}\textbf{end if}
            
  \EndFor
  \end{algorithmic}
\end{algorithm}

The value of $minsize$ of species is set by equations (5) and (6) which are as follows:
\begin{equation}
    minsize(g)=5+g/2
\end{equation}
where g represents the generation and $minsize(g)$ represents $minsize$ at generation g. During the later stages, the values of g becomes increasingly large, and hence to bound it, below equation (6) is used. 
\begin{equation}
    bound=max(10,3*D)
\end{equation}
where D represents the dimensionality of the problem.
A spanning tree is constructed similar to the standard NBC and the mean distance $\mu_{dist}$ is calculated. Next, we calculate the follow vector where every element of follow represents the number of nodes in the subtree rooted at the corresponding individual. The follow value is initially set to 1. The edges are sorted in descending order according to the fitness values of the follower individuals. Finally, for each edge, the follow value of each follower individual is added to that of a leader individual. Also, the edges in T are sorted by their Euclidean distance in descending order. Subsequently, edges satisfying the conditions are cut off. Finally, the species are obtained after cutting the edges.

\section{Proposed Methodology}
In this section, we present the opposition DE-based method for solving MMOPs. The proposed method consists of five different components which are presented in the form of Algorithms 2-8 where Algorithm 2 represents the generic framework used for solving MMOPs. Algorithms 1, and 3-8 are further components of Algorithm 2. The proposed method is tested on the IEEE CEC 2013 benchmark functions. 
\par In the next sub-section, we describe the multi-species framework used for solving MMOPs.
\subsection{Multi-Species Framework}
In a multi-species framework, the idea is to divide the randomly distributed population into sub-populations called species and perform the evolutionary process on each species independently\cite{multi-pop}. The block diagram represented by Figure 1 broadly depicts the components of the EODE Algorithm.
\begin{figure}[!htb]
\minipage{1\textwidth}
  \includegraphics[width=55ex]{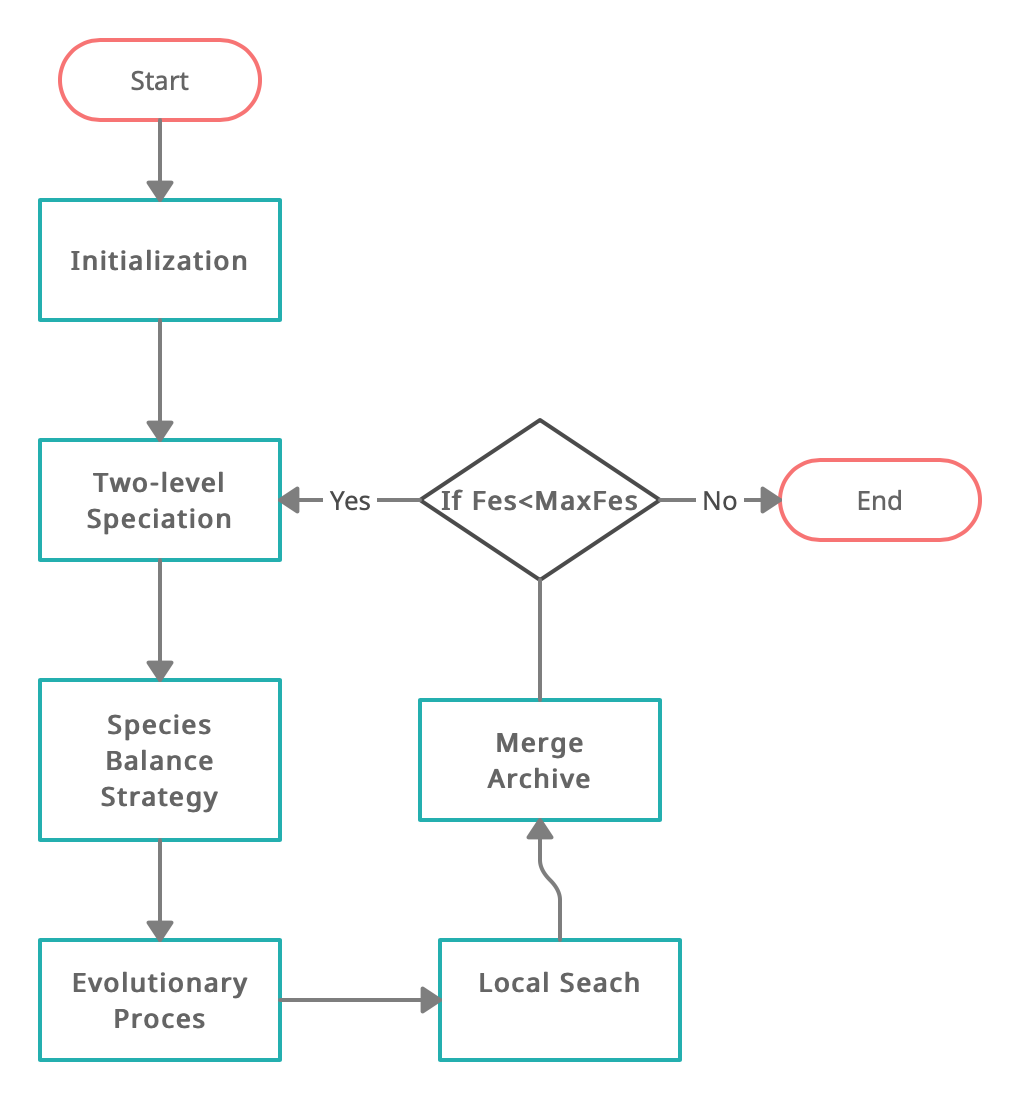}
  \newline
  \caption{Block Diagram of EODE Algorithm}
\endminipage\hfill
\end{figure}
In Figure 1, initialization is the process of randomly assigning values to the population within bounds and it corresponds to step 1 of Algorithm 2. Two-level speciation is used to divide the global population into local sub-populations or species. Two-level speciation corresponds to section 3.1.1 and step 4 of Algorithm 2. Species balance strategy is used to balance the oversized and undersized species. It corresponds to section 3.1.2 and step 5 of Algorithm 2. An evolutionary process is employed to evolve the hybrid population that has been generated by DE and ODE operations. The evolutionary process corresponds to section 3.1.3 and step 7 of Algorithm 2 while local search corresponds to section 3.1.5 and step 8 of Algorithm 2. Merge Archive procedure checks for the peaks belonging to the same species to locate multiple different peaks. It corresponds to section 3.1.6 and step 9 of Algorithm 2. Fes represents the current count of fitness evaluations while MaxFes represents the count of maximum fitness evaluations allowed. The procedures such as two-level speciation, species balance strategy, evolutionary process, local search and merge archive keep on running until the current fitness count becomes greater than or equal to the maximum count of fitness evaluations allowed.
Algorithm 2 represents the generic framework which further consists of various other algorithms.
\begin{algorithm}
\caption{EODE}\label{}
\begin{algorithmic}[1]

   \Require : Function ($f(\vec{X})$),Population Size (NP), $MaxFes$
   \State Initialize the population randomly within the bounds of the dimensions
   \State $Gen=0$,$Fes=0$,$archive=[\hspace{0.1cm}]$
   
   \While{ $Fes\leq MaxFes$}
      \State Obtain multiple species by two-level application of Algorithm 2
      \State balanceSpecies($multi$-$species$)
      \For {$species \in multi$-$species$}
        
        \State $localbest$=Modified Opposition DE($species,speciesfitness$)
        \State $bestfit$=localSearch($localbest$,$species$)  
         \State mergeArchive($archive$,$bestfit$)
      \EndFor

      \State$Gen$+=1
      \EndWhile

\end{algorithmic}
\end{algorithm}
In algorithm 2, the population is initialized randomly within the bounds specified for each dimension. The parameters $Gen$, $Fes$, $archive$ represents generation number, number of fitness evaluations, and archive, respectively. The archive is used to store optimum values. Steps 4-9 are performed until maximum fitness evaluations are reached. Algorithm 1 is invoked for creating multiple species out of the global population in step 5. After the formation of multiple species, there is a need to balance the species as each species could be representing different niches in the function landscape. To this aim, step 5 is introduced which balances the species based on the shape and size of niches. Now, for each species that is present in multi-species, steps 7-9 are carried out. In step 7, modified ODE (section 3.1.3) is applied to each species and returns the local best for that species. Local search (section 3.1.5) is a method to refine the accuracy of the best solution obtained so far. Step 8 tries to find the better solution nearby the $localbest$ achieved. It may so happen that the $localbest$ obtained is very very close to one of the optima already present in the archive and both are part of the same niche. Hence, there is a need to check the redundancy of the optimizers obtained. To this aim, step 9 is introduced. 
\par In the next section, we describe the two-level speciation procedure using the application of Algorithm 1 on the global population. 
\subsubsection{Two-level speciation}
This section corresponds to step-4 of the multi-species framework described in Algorithm 2. In this section, we describe how the global population is divided into local sub-populations using Algorithm 1.
In Algorithm 1, while using mNBC\cite{FBK-DE} for speciation, it may so happen that the root node ($e_{r}$) and follower node ($e_{f}$) belong to the same niche. In such a case, the existing mNBC will cut off the edge between $e_{r}$ and $e_{f}$ resulting in the formation of two different species even though they are part of the same species. This will lead to wastage of fitness evaluations. Hence, to improve the efficacy of mNBC, an additional check is applied in Algorithm 1. Step 12 of Algorithm 1 represents the check that checks for the presence of valley between $e_{r}$ and $e_{f}$ by using an extra fitness evaluation. Even though an extra fitness evaluation is used, it will save many unnecessary fitness evaluations. We have applied two-level speciation to capture the narrow regions of a peak. First-level speciation is performed with higher $minsize$ and hence it may associate multiple regions of a peak in a single species. To identify those close species, we employ a second level of speciation within the large species to segregate the nearby species. Second-level speciation is performed with lower $minsize$. Hence, it introduces four parameters i.e. $\varphi_{1},\varphi_{2},minsize_{1},minsize_{2}$. The parameters $\varphi_{1}, minsize_{1}$ are associated with first-level speciation and $\varphi_{2}, minsize_{2}$ are associated with second-level speciation.
\par In the next section we describe the procedure to balance the species that are obtained after two-level speciation.  

\subsubsection{Species Balance Strategy}
In this section, we describe a method to balance the species by redistributing members across the different species. Step-5 in Algorithm 2 corresponds to the species balancing strategy.
After multiple species are obtained, some species with a narrow and small basin of attraction could have many individuals and some species with a wide and large basin of attraction could have fewer individuals, hence there is a need to balance such unbalanced species. Since the niches could be of different shapes and sizes, the species balance strategy should be dynamic. In algorithm 3, initially, a minimum threshold of 10 individuals has to be present in the species which is taken care of by steps 3-4. It is done to be able to compute the covariance matrix of $t$ best individuals in the species indicated in step 6.

\begin{algorithm}
\caption{balanceSpecies}\label{}
\begin{algorithmic}[1]

   \Require{Input}: $multi$-$species$
   \For { $species \in multi$-$species$}
   \If {$speciessize\leq dim$ or $speciessize \leq$ 10}
   \State $c=max(dim-popsize,10)$
   \State Initialize $c$ individuals around the species seed
   \State  $t=max(speciessize/Gen,10)$
   \State Compute covariance matrix of $t$ best members of species
    \State  Compute variance of species using:
    $var=\sum_{i=1}^{k} eigen_{i}$
    \State Average species size$(avgsize)=\frac{NP}{\text{No of Species}}$
   \EndIf
   \If {$speciessize\textgreater\delta*avgsize$}
   \State Remove the $(speciessize-\delta*avgsize)$ worst
  individuals from the species
  \EndIf\EndFor
   \State $sortedspecies=$Sort the species in descending order of variances.
   \For {$species\in sortedspecies$}
   \If{$(species\textless\delta*avgsize)$}
   \State Generate ($\delta * avgsize$-$speciessize$) number of  individuals around
   the species seed using covariance matrix computed in step 7 and Gaussian distribution with standard deviation given by equation (3.8)
   \EndIf\EndFor

\end{algorithmic}
\end{algorithm}The variance or spread of a species is calculated by summing up the eigenvalues of the covariance matrix. The eigen values are represented using $eigen_{i}$ (step 7) where i runs from 1 to $k$ and $k$ represents the dimensionality of the problems. The average species size is computed in step 8 which considers the global population size(NP). Step 10 checks if the size of the species($speciessize$) is greater than the product of $\delta$ and the average size of the species($avgsize$) then the individuals with the least fitness values are removed from the species indicated by step 11. $\delta$ is a hyperparameter here that controls the number of new individuals generated in a species. The underbalanced species with large variance is balanced first and hence step 14. For each unbalanced species, the new individuals are generated in the vicinity of species seed using the Gaussian distribution. To this aim, Steps 15-17 are introduced.
\par In the next section we present the modified opposition DE for evolving the species that we obtained after balancing them.

\par
\subsubsection{Modified Opposition DE}
In this section, we introduce the modified opposition DE to evolve the sub-population. This procedure corresponds to step 7 mentioned in Algorithm 2 (section 3.1).
Algorithm 4 consists of the application of modified opposition DE to each species. 
\\The notations used in algorithm 4 are as below:\\
$F_{1}set$: Set to store successful $\vec{F_{1}}$ values, $F_{2}set$:Set to store successful $\vec{F_{2}}$ values, $CRset$: Set to store successful $\vec{CR}$ values, $Max_FES$: Maximum Fitness Evaluations, $FES$: Current fitness evaluations, $Gen$: Current Generation Number, $MaxGen$: Maximum Generation Number, $\vec{FB}$: First Best Member, $\vec{SB}$: Second Best Member, $\vec{TB}$: Third Best Member, $\vec{X_{rk}}$: Population Member, $pr$: Probability, NP: Population size, $\vec{U}$: Trial vector, $\vec{F}_{1}$: Mutation factor, $\vec{F}_{2}$: Mutation factor, $\vec{CR}$: Crossover rate, $JR$: Jumping Rate, $D$: Dimensions,$MIN_{j}$: Minimum in $j^{th}$ dimension , $MAX_{j}$: Maximum in $j^{th}$ dimension, $OP_{ij}$: Opposite member, $P_{ij}$: Current member , $P$: Current Population, $OP$: Opposite Population \\
Three sets are taken to store the successful values of $\vec{F_{1}}$, $\vec{F_{2}}$, and $\vec{CR}$. The successful values here mean the values which lead to the production of better offspring. $pr$ represents the probability with which the exploitation needs to be done. In early generations, more exploration of search space needs to be done while in the later phases more exploitation needs to be done. Hence, to this aim steps 7-18 are introduced. Step 9 represents the perturbation of lower degree as there is only one mutation component while step 11 represents the perturbation to be of higher degree as two mutation components are introduced. The moderate and strong exploitation are introduced by Step 16 and 18 respectively. 

\begin{algorithm}
\caption{Modified Opposition DE}\label{}
  \begin{algorithmic}[1]
\Require $species, speciesfitness$
\State $F_{1}set=\phi$, $F_{2}set=\phi$, $CRset=\phi$
\While {$FES\textless MaxFES$ and $Gen\textless MaxGen$}
\For {member$\in$ species}
\State Find first($\vec{FB}$), second($\vec{SB}$) and third($\vec{TB}$) best members in the species.
\State Randomly sample 5 members$(\vec{X_{r1}},\vec{X_{r2}},\vec{X_{r3}},\vec{X_{r4}},\vec{X_{r5}})$ from the species
\State $pr=Gen/MaxGen$
 
\If{$pr \leq$0.33}\label{algln2}
        \If {random(0,1)$\leq$ 0.75}
        \State $\vec{V}=\vec{X}_{r1}+\vec{F}_{1}*(\vec{X}_{r2}-\vec{X}_{r3})$
\Else
        \State $\vec{V}=\vec{X}_{r1}+\vec{F}_{1}*(\vec{X}_{r2}-\vec{X}_{r3})+\vec{F}_{2}*(\vec{X}_{r3}-\vec{X}_{r4})$
        \EndIf
\Else{\If{ $pr \leq$0.67}
\State Randomly select 2 members ($\vec{X_{k1}},\vec{X_{k2}}$) out of $\frac{NP}{2}$ best members
\State $\vec{V}=\vec{FB}+\vec{F}_{1}*(\vec{X}_{k1}-\vec{X}_{K2})$
\Else
\State $\vec{V}=\vec{FB}+\vec{F_{1}}*(\vec{SB}-\vec{TB})$
\EndIf}
\EndIf
\State Apply binomial crossover operation to get $\vec{U}$
\State Check for the bounds of $\vec{U}$ in each dimension using equation (5)
\State Evaluate the child fitness
\State FES+=1
\If {childfitness\textgreater parentfitness}
\State $F_{1}set\cup \vec{F_{1}}, F_{2}set\cup \vec{F_{2}}, CRset\cup \vec{CR}, $
\EndIf\EndFor
\State $JR=Gen/MaxGen$
\If {$JR\leq 1$  and $JR\textgreater 0.67$}
\State  Find the minimum, maximum bounds of the current species in each dimension.
\For{(i=0;i\textless NP;i++)}
\If {random(0,1)$\textless$ 0.33}
\For{(j=0;j\textless D;j++)}
\State $OP_{i,j}=MIN_{j}+MAX_{j}-P_{i,j}$
\EndFor
\Else
\For{(j=0;j\textless D;j++)}
\State $OP_{i,j}$=$MIN_{j}+random(0,1)*(MAX_{j}-MIN_{j}$)
\EndFor
\EndIf
\EndFor
\EndIf
\algstore{myalg} 

\end{algorithmic}
\end{algorithm}

\begin{algorithm}

  \begin{algorithmic}[1]
  \algrestore{myalg}
  
    \State Evaluate the opposite population
    \State FES+=NP
    \State Restart the candidates randomly that are stuck for 10 successive generations.
    \State Select the NP candidates with best fitness from $\{P \cup OP\}$
    \State Update $\vec{F_{1}}, \vec{F_{2}}$, and $\vec{CR}$
    \State $Gen+=1$
    \State Return the best member of the species
    \EndWhile
  \end{algorithmic}
\end{algorithm}

 Step 16 uses the candidate with the highest fitness value within that species represented by $\vec{FB}$ with a random differential vector to create the target vector. Step 18 uses the best 3 different candidates of the species. The bounds are checked using the following equation.
\begin{equation}
                 P_{i,d} =
                 \begin{cases}
 min(ub[d], 2 * lb[d] - P_{i,d}) & \text{if} (P_{i,d} \textless lb[d])  \\
 max(lb[d], 2 * ub[d] - P_{i,d}) & \text{if} (P_{i,d} \textgreater ub[d]) 
  \end{cases}
\end{equation}
where 'd' represents dimension, ub[d] and lb[d] represents upper and lower bounds in dimension 'd'. $P_{i,d}$ represents the value of $d^{th}$ dimension of $i^{th}$ member in the population.
The successful parameters such as mutation factor and crossover rate are stored in their respective sets using step 26. The jumping rate JR determines the degree of opposition applied to the species to create the opposite population. The opposite population is generated during the 33\% of the evolution process to enable better exploitation.
The key point to note here is that the mutation factors ($\vec{F1},\vec{F2}$) and crossover rate  ($\vec{CR}$) are calculated for each dimension for every generation in the population. We have used the vector representations of mutation factors and crossover rate to be able to capture the degree of perturbation in every generation individually. This helps in creating diverse offspring for better exploration.
\par
While creating the opposite population, we have introduced a tweak while creating the offspring. Steps 33 - 39 indicate the modification. Instead of creating the opposite population in the shrunken space, some amount of randomization is introduced 67\% of the time to help in the exploration of the function space. For the rest of the 33\% of time, the opposite member is calculated using step 35. During the evolution process, the population may get trapped in local optima. To avoid the trap, the population members are checked for improvement in fitness values, if they do not improve over some k (let say 10) successive generations as compared to global best member, then it is an indication of potential trap at local optima. To this end, step 46 is introduced. Step 48 is introduced at the end of every generation to update the algorithm parameters using the adaptive parameter strategy described in the next section.

\subsubsection{Adaptive Parameter Strategy}
In this section, we describe the adaptive parameter strategy to update the mutation factors and crossover rate in such a way that it adapts to the landscapes of different shapes and sizes and helps in the efficient evolution of the species. The evolution process i.e. modified ODE is discussed in section 3.1.3. The adaptive parameter strategy is used by modified ODE and it indirectly corresponds to step 7 of Algorithm 2 (section 3.1). To update the values of $\vec{F}_{1},\vec{F}_{2},\vec{CR}$, we have used a mechanism inspired from SHADE\cite{SHADE}. The complete pseudocode is given by Algorithm 5 and Algorithm 6. We have used two mutation factors ($\vec{F}_{1}$ and $\vec{F}_{2}$) to guide the population towards the optima based on the stage of evolution.  Weighted power mean is used in the calculation of $\vec{F1}$ and $\vec{CR}$ whereas weighted Lehmer mean is used in the calculation of the $\vec{F2}$. Power mean is used to interpolate between the minimum and maximum values using arithmetic mean and harmonic mean.  Lehmer mean is used to capture the non-linearity of the moving averages of the parameter values.  Weights are used to capture the improvement in influencing parameter adaptation.  Weighted Lehmer mean and weighted power mean are computed using equations (9) and (8) respectively. The $S_{\vec{F}}$ represents the set of mutation factors that produces a better offspring. $w_{k}$, computed using equation (10), represents the weight, and $\Delta{f_{k}}$ represents the fitness difference between parent and offspring. The consideration of all the dimensions while calculating the $\vec{F1}$, $\vec{F2}$, $\vec{CR}$ for next-generation helps predict the degree of perturbation to be introduced in each dimension for better exploration. $random(0,1)$ represents the random number between 0 and 1 including them. $wlm$, $wpm$ represents weighted Lehmer mean and weighted power mean respectively. $\vec{F_{old}}$ represents the $\vec{F}$ of the previous generation. $\vec{max}$, $\vec{min}$ represents the boundaries of the current species (sub-population) to which DE operations are applied. $\vec{U}$, $\vec{L}$ represents the upper and lower bounds of the objective function respectively. So, essentially the calculation of $\vec{F}$ considers the older values of $\vec{F}$, the shape and size of the basin of attraction, and the information about the stage of evolution to estimate the parameter values of the algorithm for the next generation.

\begin{equation}
Mean_{W,Power}(\vec{F}_{success})=\left(\frac{1}{\abs{S_{\vec{F}}}}\sum_{k=1}^{\abs{S_{\vec{F}}}} w_{k} *{S_{\vec{F},k}}\right)^{\frac{1}{1.5}}
\end{equation}

\begin{equation}
\operatorname{Mean}_{W, Lehmer}\left(\vec{F}_{success}\right)=\frac{\sum_{k=1}^{\abs{S_{\vec{F}}}} w_{k} \cdot S_{\vec{F}, k}^{2}}{\sum_{k=1}^{\abs{S_{\vec{F}}}} w_{k} \cdot S_{\vec{F}, k}}
\end{equation}

\begin{equation}
w_{k}=\frac{\Delta f_{k}}{\sum_{k=1}^{\abs{S_{C R}}} \Delta f_{k}}
\end{equation}
where $Mean_{W,Power}(\vec{F}_{success})$, $\operatorname{Mean}_{W, Lehmer}\left(\vec{F}_{success}\right)$ represents weighted power mean and weighted Lehmer mean. $S_{\vec{F}, k}$ represents a set of successful mutation factors while $\abs{S_{\vec{F}}}$ represents the size of the set which contains successful mutation factors. $\Delta f_{k}$ represents the fitness difference between the parent and successful offspring.

\begin{algorithm}
\caption{Adaptive Parameter Strategy($\vec{F}$)}\label{}
  \begin{algorithmic}[1]
  
  \State $wf = 0.8 + 0.2 * random(0, 1)$
   \State Compute weighted power mean for $\vec{F1}$ and weighted Lehmer mean for $\vec{F2}$ using equation (8) and (9) respectively. 

   \State $\vec{F}=0.25*\vec{F}_{old}+0.25*(\vec{max}-\vec{min})/(\vec{U}-\vec{L})+0.5*(1-\frac {FES}{MaxFes})$
   \State $\vec{F} = wf * \vec{F}_{old} + (1 - wf) * wpm$(or $wlm$)
   \end{algorithmic}
\end{algorithm}
\begin{algorithm}
\caption{Adaptive Parameter Strategy($\vec{CR}$)}\label{}
  \begin{algorithmic}[1]
  
  \State $wf = 0.9 + 0.1 * random(0, 1)$
   \State Compute weighted power mean using equation (8).

   \State $\vec{CR} = wf * \vec{CR_{old}} + (1 - wf) * wpm$
   \end{algorithmic}
\end{algorithm}
 The adaptive parameter strategy for mutation factors is described in Algorithm 5. Algorithm 6 introduces an adaptive parameter strategy for crossover rate. In the next section, we define the local search method to improve the best solution obtained after applying modified ODE in the hope to further improve its fitness.

\subsubsection{Local Search}
Local search methods are generally used to refine the accuracies of the obtained solutions. It is being used in step 8 of Algorithm 2. Our proposed local search method is described in Algorithm 7. Essentially, we generate some members in the vicinity of the best candidate obtained so far in the species in the hope to find a further better candidate. We initialize variances ($vars$) using step 3 described in algorithm 7. The key idea is to keep incrementing the variances ($vars$) if the better offspring are not found in the nearby region of the best individual of the species. dim represents the dimensionality of the problem. $dirvec$, computed in step 18, can be visualized as a direction vector and it is used to perform a guided search nearby the $localbest$ candidate. The degree of the shift from the $localbest$ is continuously updated in the hope of finding a better solution by incrementing the variances. In Algorithm 7, Gaussian distribution has been used to generate offsprings around the $localbest$ in the current species. For efficient search in the nearby region of the $localbest$ candidate, we generate offspring by differential vector perturbation and estimating the distribution of the handful of better candidates in the species.  Mean and standard deviation are computed using (9), (10). 
\begin{equation}
\mu_{i}^{d}=\frac{1}{M} \sum_{j=1}^{M} X_{j}^{d}
\end{equation}
\begin{equation}
\delta_{i}^{d}=\sqrt{\frac{1}{M-1} \sum_{j=1}^{M}\left(X_{j}^{d}-\mu_{i}^{d}\right)^{2}}
\end{equation}
where $\mu_{i}=\left[\mu_{i}^{1}, \ldots, \mu_{i}^{d}, \ldots, \mu_{i}^{D}\right]$ and $\delta_{i}=$
$\left[\delta_{i}^{1}, \ldots, \delta_{i}^{d}, \ldots, \delta_{i}^{D}\right](1\leq i \leq s)$ are, respectively, the
mean and standard deviation (std) vectors of the $i^{th}$ niche, $\boldsymbol{X}_{j}=\left[X_{j}^{1}, \ldots, X_{j}^{d}, \ldots, X_{j}^{D}\right]$ is the $j^{th}$ individual in the $i^{th}$ niche and $D$ is the dimension size of the multimodal problem.

\begin{algorithm}
  \caption{localSearch}\label{localSearch}
  \begin{algorithmic}[1]
   
      \Require :$localbest$,$species$
      \State $dirvec$=None
      \State k=10
      \State $vars=random(0.001,0.01)$
      \While{(k)}
      \State $mean=localbest$
     \If {$dirvec$ is not None and $random(0,1)\leq$ 0.5}
      \State offspring[k]=$localbest$+$vars$*$dirvec$
      \Else
      \State $mbest$=max(speciessize/4,10)
      \State Find the covariance matrix of the $mbest$ candidates of species using equation (11)
      \If {dim$\textgreater$ 1}
      \State offspring[k]=Apply multivariate normal distribution with mean as $mean$ and covariance matrix as computed in step 10
      \Else
      \State offspring[k]=Apply univariate normal distribution with mean as $mean$ and standard deviation is as computed using equation (10)
      \EndIf
      \EndIf
      \If {$\vec{O}_{fitness}$ \textgreater $\vec{X}_{bestfitness}$}
      \State $dirvec$=offspring[k]-$localbest$
      \State $localbest$=offspring[k]
      \Else
      \For {d in dim}
      \State $vars[d]$+=$random(0.001,0.01)$
      \EndFor
      \EndIf
      \State k=k-1
      \EndWhile
      \State Return the $localbest$
  \end{algorithmic}
\end{algorithm}
 
 The covariance matrix is used to capture the behavior of variances of each dimension with respect to the other dimensions. Covariance between two variables of dimension D is computed as follows:
 \begin{equation}
\mathrm{COV_{X,Y}}=\frac{\sum_{i=1}^{D}\left(X_{i}-\bar{x}\right)\left(Y_{i}-\bar{y}\right)}{D-1}
\end{equation}
where $\bar{x},\bar{y}$ represents X mean and Y mean.
 $\vec{O}_{fitness}, \vec{X}_{bestfitness}$ represents the offspring fitness and member with the best fitness respectively. In the next section, the mergeArchive procedure is discussed that aims at removing the peaks belonging to the same niche.

\subsubsection{Merge Archive}
In this section, we describe the procedure to deal with duplicate peaks in the $archive$. This section corresponds to step 9 of Algorithm 2.
Since we are storing the peaks in an $archive$, it is important to store the peaks only. Otherwise, we may end up storing the entire candidates of a sub-population (species) and subsequently the entire population. This will lead to the consumption of more computational resources since the $archive$ will contain huge points. Hence,
given two individuals, it is essential to check whether they belong to similar species or different species. It is critical to algorithm performance in locating all the global optima. In algorithm 8, the inputs are $localbest$, $archive$ and it needs to be checked whether the $localbest$ is a new species seed or an old species seed. The nearest peak to the given local best is found out, and the midpoint between them is calculated. If the midpoint has lower fitness than the nearest peak and $localbest$, it indicates the presence of a valley between them. Hence, local best can be considered as the species seed of new species, and it is added to the $archive$.
In the next section, we describe the experiments performed and the results obtained.
\begin{algorithm}
  \caption{mergeArchive}\label{mergeArchive}
  \begin{algorithmic}[1]
    
      \Require {$archive,localbest$}
      \State Find the nearest peak to $localbest$ present in the $archive$
      \State Find the mid-point between $localbest$ and nearest peak
      \If {(midpointfitness \textgreater $localbest$ and midpointfitness \textgreater nearest peak)}
      \State Add $localbest$ to the $archive$
      \Else
       \State Replace the better fit individual between $localbest$ and nearest peak with mid-point. 
      \EndIf
  \end{algorithmic}
\end{algorithm}

\subsection{Illustrative Example}
In order to explain the components of EODE, an example is presented in this section. We have considered low dimensional (2-D) composition function $F_{11}$.

\begin{figure}[!htb]
\minipage{0.5\textwidth}
  \includegraphics[width=\linewidth]{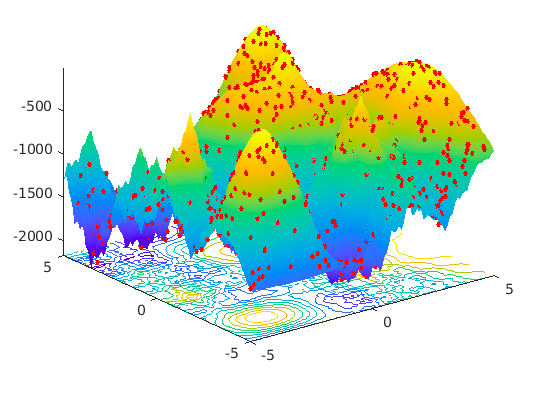}
  \caption{Initialization}\label{fig:F1}
\endminipage\hfill
\minipage{0.5\textwidth}
  \includegraphics[width=\linewidth]{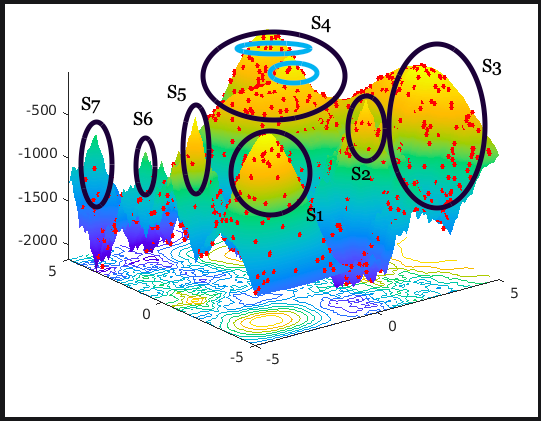}
  \caption{Niching and Balancing}\label{fig:F1}
\endminipage\hfill

\end{figure}
\begin{figure}[!htb]
\minipage{0.5\textwidth}
  \includegraphics[width=\linewidth]{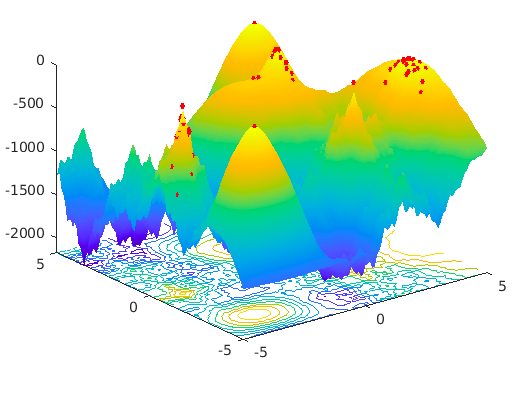}
  \caption{ODE and Local Search}\label{fig:F2}
\endminipage\hfill
\end{figure}
As it can be seen from Figure 2, the population is randomly initialized in the function space. Now we will try to map the steps given in Algorithm 2 to the processes that happen in Fig. 2-4. 
\begin{enumerate}
    \item Initially, the randomly distributed population needs to be divided into sub-populations that represent regions around the peaks. The multiple species marked as black circles in Fig 3 are shown as S1-S7. In species S4, the blue circles represent two species that got located as a single species S4. To avoid such merging, two-level speciation is introduced. To this aim, step 4 (Algorithm 2) is introduced.
    \item As the multiple species are formed, some species may have huge population size and some may have small population size as can be seen from Figure 3. Hence, the species balance strategy is introduced that aims to generate offspring using Gaussian distribution. Also, the species balance strategy is discussed in Algorithm 3 which maps to step 5 in Algorithm 2.
    \item Modified ODE (section 3.1.3) acts as an evolutionary process that tries to further converge the sub-populations towards the peak they represent. The species are represented in Figure 4 as P1-P8. Further, it is more effective to converge towards global optima rather than local optima. It can be seen in Figure 4 in which P7 and P8 are local optima. This process maps to step 7 in Algorithm 2.
    \item After the evolution process completes, the population gets converged near the optima but may not be able to reach the optima accurately as can be seen in Figure 4. Hence, we proposed a local search method to search in the vicinity of the best individual of the species as the actual peak would be nearby the best value in the species. The local search method is described in Algorithm 7 and it maps to step 8 of Algorithm 2.
    \item Since the speciation process is repeated after every certain number of generations, the best solutions returned by the local search method could be part of the same species that has been explored in the previous generations. Hence, there is a need to detect the best individuals that belong to the same species and they should not be added to the archive. Therefore, Algorithm 8 is introduced that maps to step 9 of Algorithm 2.

\end{enumerate}

\begin{table}

\renewcommand{\arraystretch}{1.2}
\caption{\label{tab:table-name}Information of the benchmark problems and the population size}
\begin{tabular}{llllll|l}
\hline Index & Function & NKP & Peak height & $r$ & MaxFEs & $N P$ \\
\hline 1 & $F_{1}(1 D)$ & 2 & $200.0$ & $0.01$ & $5.0 \mathrm{E}+4$ & 250 \\
2 & $F_{2}(1 D)$ & 5 & $1.0$ & $0.01$ & $5.0 \mathrm{E}+4$ & 250 \\
3 & $F_{3}(1 D)$ & 1 & $1.0$ & $0.01$ & $5.0 \mathrm{E}+4$ & 250 \\
4 & $F_{4}(2 D)$ & 4 & $200.0$ & $0.01$ & $5.0 \mathrm{E}+4$ & 250 \\
5 & $F_{5}(2 D)$ & 2 & $1.03163$ & $0.5$ & $5.0 \mathrm{E}+4$ & 250 \\
6 & $F_{6}(2 D)$ & 18 & $186.731$ & $0.5$ & $2.0 \mathrm{E}+5$ & 2000 \\
7 & $F_{7}(2 D)$ & 36 & $1.0$ & $0.2$ & $2.0 \mathrm{E}+5$ & 2000 \\
8 & $F_{6}(3 D)$ & 81 & $2709.0935$ & $0.5$ & $4.0 \mathrm{E}+5$ & 3000 \\
9 & $F_{7}(3 D)$ & 216 & $1.0$ & $0.2$ & $4.0 \mathrm{E}+5$ & 4000 \\
10 & $F_{8}(2 D)$ & 12 & $-2.0$ & $0.01$ & $2.0 \mathrm{E}+5$ & 1000 \\
11 & $F_{9}(2 D)$ & 6 & 0 & $0.01$ & $2.0 \mathrm{E}+5$ & 1000 \\
12 & $F_{10}(2 D)$ & 8 & 0 & $0.01$ & $2.0 \mathrm{E}+5$ & 1000 \\
13 & $F_{11}(2 D)$ & 6 & 0 & $0.01$ & $2.0 \mathrm{E}+5$ & 1000 \\
14 & $F_{11}(3 D)$ & 6 & 0 & $0.01$ & $4.0 \mathrm{E}+5$ & 1000 \\
15 & $F_{12}(3 D)$ & 8 & 0 & $0.01$ & $4.0 \mathrm{E}+5$ & 1000 \\
16 & $F_{11}(5 D)$ & 6 & 0 & $0.01$ & $4.0 \mathrm{E}+5$ & 1000 \\
17 & $F_{12}(5 D)$ & 8 & 0 & $0.01$ & $4.0 \mathrm{E}+5$ & 2000 \\
18 & $F_{11}(10 D)$ & 6 & 0 & $0.01$ & $4.0 \mathrm{E}+5$ & 1000 \\
19 & $F_{12}(10 D)$ & 8 & 0 & $0.01$ & $4.0 \mathrm{E}+5$ & 1000 \\
20 & $F_{12}(20 D)$ & 8 & 0 & $0.01$ & $4.0 \mathrm{E}+5$ & 800 \\
\hline
\end{tabular}

\end{table}

\begin{table}

\centering

\caption{Parameters in EODE}
\begin{tabular}{l l}
\hline Parameters & Values \\
$\varphi_{1}$ & $1$ \\
$\varphi_{2}$ & $1$ \\
$minsize_{1}$ & $-1$ \\
$minsize_{2}$ & $5$ \\
$\delta$ & $1.0$ \\
$F1$&$(0,1)$\\
$F2$&$(0,1)$\\
$CR$&$(0,1)$\\

$MaxGen$ & $40(D<=10)$ \\
& $60(D \textgreater 10)$ \\
\hline

\end{tabular}

\end{table}
\section{Experiments and Results}
We have performed the experiments on a computer system with RAM 8GB, 1.8GHz CPU and MacOS 11 operating system.
In this section, EODE is independently run 50 times for each function. The algorithmic results are computed in five levels of accuracy $\epsilon$ = \{$1\mathrm{e}{-1}, 1\mathrm{e}{-2}, 1\mathrm{e}{-3}, 1\mathrm{e}{-4}, 1\mathrm{e}{-5}$\}.

\begin{table}
\renewcommand{\arraystretch}{1.5}

\centering\caption{Results on accuracy levels 1e-1,1e-2,1e-3,1e-4,1e-5}
\begin{center}
\begin{tabular}{|l|l|l|l|l|l|}
\hline

  & 1E-1 & 1E-2 & 1E-3 & 1E-4 & 1E-5 \\
\hline
 Functions & (PR,SR) & (PR,SR) & (PR,SR) & (PR,SR) & (PR,SR) \\
 \hline
 F1(1D) & (1,1)& (1,1)& (1,1)& (1,1)& (1,1)\\
 F2(1D) & (1,1)& (1,1)& (1,1)& (1,1)& (1,1)\\
 F3(1D) & (1,1)& (1,1)& (1,1)& (1,1)& (1,1)\\
 F4(2D) & (1,1)& (1,1)& (1,1)& (1,1)& (1,1)\\
 F5(2D) & (1,1)& (1,1)& (1,1)& (1,1)& (1,1)\\
 F6(2D) & (1,1)& (0.995,0.9)& (0.995,0.9)& (0.995,0.9)& (0.9,0.822)\\
 F7(2D) & (0.805,0)& (0.805,0)& (0.805,0)& (0.805,0)& (0.805,0)\\
 F6(3D) & (0.886,0)& (0.852,0)& (0.852,0)& (0.845,0)& (0.832,0)\\
 F7(3D) & (0.509,0)& (0.505,0)& (0.505,0)& (0.505,0)& (0.442,0)\\
 F8(2D) & (1,1)& (1,1)& (1,1)& (1,1)& (1,1)\\
 F9(2D) & (1,1)& (1,1)& (1,1)& (1,1)& (1,1)\\
 F10(2D) & (0.975,0.8)& (0.975,0.8)& (0.975,0.8)& (0.975,0.8)& (0.975,0.8)\\
 F11(2D) & (1,1)& (1,1)& (1,1)& (1,1)& (1,1)\\
 F11(3D) & (0.8,0)& (0.8,0)& (0.8,0)& (0.8,0)& (0.8,0)\\
 F12(3D) & (0.8,0)& (0.8,0)& (0.8,0)& (0.8,0)& (0.77,0)\\
 F11(5D) & (0.733,0)& (0.730,0)& (0.730,0)& (0.730,0)& (0.728,0)\\
 F12(5D) & (0.7,0)& (0.7,0)& (0.7,0)& (0.684,0)& (0.684,0)\\
 F11(10D) & (0.7,0)& (0.7,0)& (0.7,0)& (0.684,0)& (0.684,0)\\
 F12(10D) & (0.525,0)& (0.525,0)& (0.525,0)& (0.520,0)& (0.505,0)\\
 F12(20D) & (0.25,0)&  (0.25,0)&  (0.25,0)&  (0.25,0)&  (0.25,0)\\
\hline
\end{tabular}
\end{center}
\end{table}
 For different problems, the value of NP is shown in Table 1.  To decide upon the values of NP, the algorithm is run for 50 times with different population sizes for each function and that NP is chosen which provides best value of average PR over 50 runs for the given function. The results are depicted by the Table 4. The $NP_{1},NP_{2},NP_{3},NP_{4}$ and $NP_{5}$ represents different population sizes for each function and the $PR_{1},PR_{2},PR_{3},PR_{4}$ and $PR_{5}$ represents the corresponding average PR values for the given population sizes $NP_{i}$.
 \begin{table}
\renewcommand{\arraystretch}{1.5}

\centering\caption{PR values on different population sizes}
\begin{center}
\begin{tabular}{|l|l|l|l|l|l|l|l|l|l|l|}

\hline

 Functions & $NP_{1}$ &$PR_{1}$ & $NP_{2}$ &$PR_{2}$& $NP_{3}$&$PR_{3}$& $NP_{4}$ &$PR_{4}$ & $NP_{5}$ &$PR_{5}$\\
 \hline
 F1(1D) & 50&1&100&1&250&1&500&1&1000&1\\
 F2(1D) & 50&1&100&1&250&1&500&1&1000&1\\
 F3(1D) & 50&1&100&1&250&1&500&1&1000&1\\
 F4(2D) & 50&0.6&100&0.93&250&1&500&1&1000&1\\
 F5(2D) & 50&1&100&1&250&1&500&1&1000&1\\
 F6(2D) & 500&0.966&1000&0.968&2000&0.995&3000&0.985&5000&0.911\\
 F7(2D) & 500&0.466&1000&0.695&2000&0.805&3000&0.797&5000&0.659\\
 F6(3D) & 500&0.506&1000&793&2000&0.829&3000&0.845&8000&0.817\\
 F7(3D) & 1000&0.264&2000&0.479&4000&0.505&8000&0.493&10000&0.427\\
 F8(2D) & 250&0.65&500&0.895&1000&1&2000&0.969&3000&0.912\\
 F9(2D) & 100&0.633&250&0.998&500&1&1000&1&2000&1\\
 F10(2D) & 100&0.475&250&0.781&500&0.919&1000&0.975&2000&0.958\\
 F11(2D) & 100&0.366&250&0.766&500&0.933&1000&1&2000&0.993\\
 F11(3D) & 100&0.366&250&0.688&500&0.788&1000&0.8&2000&0.697\\
 F12(3D) & 100&0.4&250&0.795&500&0.639&1000&0.8&2000&0.795\\
 F11(5D) & 500&0.466& 1000&0.730& 2000&0.719& 5000&0.685& 8000&0.644\\
 F12(5D) & 500&0.433& 1000&0.684& 2000&0.598& 5000&0.515& 8000&0.489\\
 F11(10D) & 500&0.466& 1000&0.684& 2000&0.681& 5000&0.589& 8000&0.512\\
 F12(10D) & 500&0.375& 1000&0.520& 2000&0.505& 5000&0.318& 8000&0.305\\
 F12(20D) &500&0.175&800&0.250&1000&0.218&2000&0.198&5000&0.195\\
\hline
\end{tabular}
\end{center}
\end{table}
 
 The parameters of EODE are listed in Table 2. $\varphi_{1}$, $\varphi_{2}$ are used for creating multiple species out of the global population by applying mNBC twice and their values are set as 1, 1 respectively. $minsize_{1}$, $minsize_{2}$ are parameters used in two-level application of mNBC and their values are set as -1, 5 respectively. We define the $minsize_{1}$ as -1 to adapt the first level species as per the dimension and generation of the population. $minsize_{2}$ is set to be 5 to identify the peaks that have narrow basins of attraction as the species size would be very low in such regions.  $\delta$ is used in the species balance strategy and it is set to 1. 

 The mutation factors ($\vec{F1}, \vec{F2}$) and crossover rate ($\vec{CR}$) are initially chosen from the closed interval [0,1] after which the algorithm learns to adapt their values. $MaxGen$ represents the number of generations for which the ODE needs to run for the given species.  The experimental results of EODE for all benchmark problems are listed in Table 3  at all five accuracy levels. These results show that EODE is very stable. In addition, EODE finds all peaks on the simple functions and most of the peaks in the low-dimensional composition problems. EODE can find more than 80\% of all peaks on the functions containing a large number of global peaks, except for F7(3D).

\subsection{Comparison with other algorithms}
We have compared the results of various algorithms (including EODE) in this section. We have compared the results at $\epsilon = 1\mathrm{e}{-4}$, which are commonly adopted in \cite{Wangg} and \cite{Yangg}. 
To have an extensive comparison, we have evaluated the performance of EODE and compared it with 15 popular  algorithms such as CDE \cite{crowding}, SDE\cite{DE5}, NCDE, NSDE \cite{Qu-DE}, MOMMOP \cite{MOMMOP}, LoICDE, LoISDE \cite{ada2}, PNPCDE \cite{pnpcde}, LIPS \cite{Qu}, and the
 recently proposed algorithms of $DE_{cl}$ \cite{DEcl}, LMCEDA, LMSEDA \cite{EDA}, FBK-DE\cite{FBK-DE}, LBPADE\cite{Zhao}, MaHDE\cite{MaHDE}. Table 4, 5, 6 shows the different PRs and SRs at the accuracy level $\epsilon = 1\mathrm{e}{-4}$, where the bottom row “bprs” represents the number of functions that achieved best PR values for these algorithms. The algorithm(s) with the best PR value for a given function is marked in bold. All the algorithms used their own default population sizes. The results of the compared algorithms have been referred from their respective papers.
From Table 4, 5, 6, it is clear that EODE obtains the best PR values for most of the functions among the compared algorithms. The detailed analysis is given below.\\
1) For functions 1-5, most of the algorithms including EODE can find all global optimal solutions.\\
2) For functions 6-9 which consists of a large number of global peaks, EODE does not perform equally well as compared to all the algorithms. However, it achieved better results as compared to many algorithms. For problem 6, the difference is of order 0.005 which is very small. Also, for the $10^{th}$ problem, EODE can find
all the optima.\\
3) For functions 11,12,13 and 15 that are low-dimensional composition functions, EODE obtained the best results. Although on the $14^{th}$ function EODE does not produce optimal results, the difference between EODE and the best algorithm is small (around 10\%). It is important to note that EODE and FBK-DE  are the two algorithms that are able to find all global optima on the $13^{th}$ function while the other algorithms do not.\\ 
4) For the functions $16^{th}$–$20^{th}$ that are composition functions having relatively high dimensionality, EODE achieves the best results for all functions except the $20^{th}$ function. It is worth noting that on the 5-D problems, the result obtained by EODE exceeds 5\% of the best result obtained by other algorithms. For function 18, EODE performed a little better than the best algorithm while for the $19^{th}$ function it performed equivalently well.

\begin{sidewaystable}
\renewcommand{\arraystretch}{1.2}
\sidewaystablefn%
\begin{center}
\begin{minipage}{\textheight}
\caption{Comparison of EODE with other algorithms on accuracy level 1e-4}\label{tab3}

\begin{tabular}{|l|l|l|l|l|l|l|l|l|l|l|l|l|l}
\hline Function & \multicolumn{2}{|c|} {EODE} & \multicolumn{2}{c|} { CDE }& \multicolumn{2}{c|} { SDE }& \multicolumn{2}{c|} { NCDE }& \multicolumn{2}{c|} { NSDE }& \multicolumn{2}{c|} { MOMMOP }\\
  Index & PR & SR & PR & SR& PR & SR& PR & SR& PR & SR& PR & SR \\
  \hline
 1&\textbf{1} & 1 & \textbf{1} & 1 & 0.657& 0.373 & \textbf{1}& 1 & \textbf{1}& 1 & \textbf{1}& 1  \\
2 &\textbf{1}& 1 & \textbf{1} & 1 & 0.737& 0.529 & \textbf{1}& 1 &  0.776 & 0.667&\textbf{1} &1\\
3 & \textbf{1} & 1 & \textbf{1} & 1& \textbf{1} & 1& \textbf{1} & 1& \textbf{1} & 1& \textbf{1}&1 \\
4 & \textbf{1} & 1 &\textbf{1} & 1& 0.284& 0 & \textbf{1}& 1 &  0.240 & 0& \textbf{1}&1 \\
5 & \textbf{1} & 1 & \textbf{1} & 1&  0.922& 0.843 & \textbf{1}& 1 & 0.745& 0.490 & \textbf{1}&1 \\
6 & 0.995 & 0.9 & \textbf{1} & 1&  0.056& 0 & 0.305& 0 & 0.056& 0 & \textbf{1}&1 \\
7 & 0.805 & 0 & 0.861 & 0&  0.054& 0 & 0.873& 0 & 0.053& 0 & \textbf{1}&1 \\
8 & 0.845 & 0 & 0 & 0&  0.015& 0 & 0.001& 0 & 0.013& 0 & \textbf{1}&1 \\
9 & 0.505 & 0 & 0.474 & 0& 0.011& 0 & 0.461& 0 & 0.006& 0 & \textbf{1}&1 \\
10 & \textbf{1} & 1 & \textbf{1} & 1&  0.147& 0 & 0.989& 0.863 & 0.098& 0 & \textbf{1}&1 \\
11 & \textbf{1} & 1 & 0.330 & 0&  0.314& 0 & 0.729& 0.059 & 0.248& 0 & 0.716&0.020 \\
12 & \textbf{0.975} & 0.8 & 0.002 & 0&  0.208& 0 & 0.252& 0 & 0.135& 0 & 0.939&0.549 \\
13 & \textbf{1} & 1 & 0.141 & 0&  0.297& 0 & 0.667& 0 & 0.225& 0 & 0.667&0 \\
14 & 0.8 & 0 & 0.026 & 0&  0.216& 0 & 0.667& 0 & 0.190& 0 & 0.667&0 \\
15 & \textbf{0.8} & 0 & 0.005 & 0&  0.108& 0 & 0.319& 0 & 0.125& 0 & 0.618&0 \\
16 & \textbf{0.730} & 0 & 0 & 0&  0.108& 0 & 0.667& 0 & 0.170& 0 & 0.650&0 \\
17 & \textbf{0.684} & 0 & 0 & 0&  0.076& 0 & 0.250& 0 & 0.108& 0 & 0.505&0 \\
18 & \textbf{0.684} & 0 & 0.167 & 0&  0.026& 0 & 0.500& 0 & 0.163& 0 & 0.497&0 \\
19 & \textbf{0.520} & 0 & 0 & 0&  0.105& 0 & 0.348& 0 & 0.098& 0 & 0.223&0 \\
20 & 0.250 & 0 & 0 & 0&  0& 0 & 0.250& 0 & 0.123& 0 & 0.125&0 \\
\hline bprs & \multicolumn{2}{|l|} {\textbf{14}} & \multicolumn{2}{l|} { 7 }& \multicolumn{2}{l|} { 1 }& \multicolumn{2}{l|} { 5 }& \multicolumn{2}{l|} { 2 }& \multicolumn{2}{l|} { 10 }\\
\hline
\end{tabular}

\end{minipage}
\end{center}
\end{sidewaystable}

\begin{sidewaystable}
\renewcommand{\arraystretch}{1.2}
\sidewaystablefn%
\begin{center}
\begin{minipage}{\textheight}
\caption{Comparison of EODE with other algorithms on accuracy level 1e-4}\label{tab3}
\begin{tabular}{|l|l|l|l|l|l|l|l|l|l|l|l|l|l}
\hline Function & \multicolumn{2}{|c|} { EODE } & \multicolumn{2}{|c|} { LoICDE }& \multicolumn{2}{|c|} { LoISDE } & \multicolumn{2}{|c|} {PNPCDE } & \multicolumn{2}{|c|} {LIPS } & \multicolumn{2}{|c|} {$DE_{cl}$ } \\
Index & PR & SR & PR & SR& PR & SR& PR & SR& PR & SR& PR & SR \\
\hline
1 &  \textbf{1}&1 & \textbf{1}&1  & \textbf{1}& 1 &\textbf{1}& 1 & 0.833& 0.686 & \textbf{1}& 1   \\
2 &\textbf{1} &1&\textbf{1}& 1 & 0.235 & 0.039 & \textbf{1}& 1 & {1}& 1 &  \textbf{1} & 1\\
3 &\textbf{1}& 1& \textbf{1} & 1 &\textbf{1} & 1& \textbf{1} &1& 0.961& 0.961 & \textbf{1}& 1  \\
4 & \textbf{1}&1& 0.975 & 0.902 & 0.250 & 0& \textbf{1}& 1 & 0.990& 0.961 &  \textbf{1} & 1 \\
5 & \textbf{1} & 1 & \textbf{1} & 1&  0.667& 0.333 & \textbf{1}& 1 & \textbf{1}& 1 & \textbf{1}&1 \\
6 & 0.995 & 0.9 & \textbf{1} & 1&  0.056& 0 &  0.537 & 0& 0.246 &0 & 0.942&0.340 \\
7 &  0.805 & 0 &0.705&0.02&0.029&0&0.874&0&	0.4&0&0.986&0.64 \\
8 & 0.845 & 0 & 0& 0&	0.012&0&0&0&	0.084&0&	0.999&0.9\\
9 & 0.505 & 0 & 0.187&0&	0.005&0& 0.472&0&	0.104&0	&0.726&0 \\
10 & \textbf{1} & 1 &\textbf{1}&1&	0.083&0&\textbf{1}&1&	0.748&0&	\textbf{1}&1 \\
11 & \textbf{1} & 1 &0.66&0&	0.167&0&0.66&0&	0.974&0.843&	0.667&0 \\
12 & \textbf{0.975} & 0.8 & 0.495&0&	0.125&0&0&0&	0.574&0&	0.943&0.58 \\
13 & \textbf{1} & 1 & 0.51&0&	0.167&0&0.461&0&	0.794&0.176&	0.667&0 \\
14 & 0.8 & 0 & 0.657&0&	0.167&0&0.592&0&	0.644&0&	0.667&0 \\
15 & \textbf{0.8} & 0 & 0.299&0&	0.125&0&0.258&0	&0.336&0&	0.623&0 \\
16 & \textbf{0.730} & 0 &0.559&0&	0.167&0&0&0&	0.304&0&	0.667&0 \\
17 & \textbf{0.684} & 0 & 0.223&0&	0.076&0&0&0&	0.162&0&	0.42&0 \\
18 &\textbf{0.684} & 0 & 0.219&0&	0.157&0&0.147&0&	0.098&0&	0.667&0 \\
19 & \textbf{0.520} & 0 & 0.037&0&	0.027&0&0&0&	0&0&	0.357&0 \\
20 & 0.250 & 0 & 0.123&0&	0.088&0&0&0&	0&0&	0.212&0 \\
\hline bprs & \multicolumn{2}{|c|} {\textbf{14}} & \multicolumn{2}{c|} { 6}& \multicolumn{2}{c|} { 2 }& \multicolumn{2}{c|} { 6 }& \multicolumn{2}{c|} { 2 }& \multicolumn{2}{c|} { 6 }\\

\hline
\end{tabular}

\end{minipage}
\end{center}
\end{sidewaystable}

\begin{sidewaystable}
\renewcommand{\arraystretch}{1.2}
\sidewaystablefn%
\begin{center}
\begin{minipage}{\textheight}
\caption{Comparison of EODE with other algorithms on accuracy level 1e-4}\label{tab3}
\begin{tabular}{|l|l|l|l|l|l|l|l|l|l|l|l|l|l}
\hline Function & \multicolumn{2}{|c|} { EODE } & \multicolumn{2}{|c|} { LMCEDA }& \multicolumn{2}{c|} {LMSEDA }&
\multicolumn{2}{c|} { FBK-DE }& \multicolumn{2}{c|} {LBPADE }& \multicolumn{2}{c|} {MaHDE } \\
 Index & PR & SR & PR & SR& PR & SR& PR & SR& PR & SR& PR & SR \\
 \hline
1 & \textbf{1}& 1&\textbf{1} & 1 & \textbf{1} & 1 &\textbf{1} & 1 &\textbf{1} & 1&\textbf{1}& 1    \\
2 & \textbf{1}& 1&\textbf{1} & 1 & \textbf{1} & 1 &\textbf{1} & 1 &\textbf{1} & 1&\textbf{1} & 1 \\
3 & \textbf{1}& 1&\textbf{1} & 1 & \textbf{1}& 1 &\textbf{1} & 1 &\textbf{1}& 1&\textbf{1} & 1   \\
4 & \textbf{1}& 1&\textbf{1} & 1 & \textbf{1} & 1 &\textbf{1} & 1 &\textbf{1}& 1&\textbf{1} & 1 \\
5 &  \textbf{1}& 1&\textbf{1} & 1 & \textbf{1} & 1 &\textbf{1} & 1 &\textbf{1} & 1&\textbf{1} & 1  \\
6 & 0.995 & 0.9 & 0.99&0.843&	0.972&0.588&0.990&0.820&\textbf{1}&1&\textbf{1}&1 \\
7 &  0.805 & 0 &0.734&0&	0.673&0&0.813&0&0.889&0&0.804&0 \\
8 & 0.845 & 0 & 0.367&0&	0.613&0&0.824&0&0.575&0&0.983&0.291\\
9 & 0.505 & 0 & 0.284&0&	0.248&0& 0.425&0&0.476&0&0.351&0\\
10 & \textbf{1} & 1 &\textbf{1}&1&	0.998&0.98&\textbf{1}&1&\textbf{1}&1&\textbf{1}&1 \\
11 & \textbf{1} & 1 &0.667&0&	0.892&0.392& \textbf{1}&1&0.674&0&0.725&0.078\\
12 & \textbf{0.975} & 0.8 & 0.75&0&	0.99&0.922&0.935&0.480&0.750&0&0.650&0 \\
13 & \textbf{1} & 1 & 0.667&0&	0.667&0&1&1&0.667&0&0.667&0\\
14 & 0.8 & 0 & 0.667&0&	0.667&0&\textbf{0.907}&0.460&0.667&0&0.667&0 \\
15 & \textbf{0.8} & 0 & 0.696&0&	0.738&0&0.730&0&0.654&0&0.648&0 \\
16 & \textbf{0.730} & 0&0.667&0&	0.667&0 &0.707&0&0.667&0&0.667&0 \\
17 & \textbf{0.684} & 0 &0.456&0&	0.62&0&0.630&0&0.532&0&0.352&0 \\
18 &\textbf{0.684} & 0 & 0.657&0&	0.66&0&0.667&0&0.667&0&0.663&0\\
19 & \textbf{0.520} & 0 & 0.451&0&	0.458&0&\textbf{0.520}&0&0.475&0&0.455&0 \\
20 & 0.250 & 0 & 0.059&0&	0.248&0&\textbf{0.450}&0&0.275&0& 0.250&0\\
\hline bprs & \multicolumn{2}{|c|} {\textbf{14}} & \multicolumn{2}{c|} { 6 }& \multicolumn{2}{c|} { 5 }& \multicolumn{2}{c|} { 10 }& \multicolumn{2}{c|} { 7 }& \multicolumn{2}{c|} { 7 }\\
\hline
\end{tabular}

\end{minipage}
\end{center}
\end{sidewaystable}

\subsection{Different Components of EODE}
In this section, we have discussed the different components embedded in EODE and their effect on the algorithm. We have also discussed the mutation operators that are used based on evolution stages. This section also contains a brief discussion about the effect of $\varphi_{1}$ and $\varphi_{2}$ that are key parameters in deciding the formation of the total number of species. We briefly describe the effects of JR and the convergence of population towards global optima.  
\subsubsection{Different Mutation Operators}
We have compared the results of the algorithms using different mutation operators one at a time. The balancing ability of the algorithm is shown by comparing the four different algorithms denoted as EODE, EODE-r, EODE-b, EODE-rb. DE/rand/1 and DE/rand/2 are used as mutation operators in EODE-r to evolve the species. Besides, EODE-b uses both DE/best/1 and DE/best/2 while EODE-rb uses DE/rand/1, DE/rand/2, DE/best/1 and DE/best/2 for mutation, which can be referred from \cite{mutation}.

\begin{table}
\renewcommand{\arraystretch}{1.2}
\centering
\caption{\label{tab:table-name}Experimental results of different mutation operators on the benchmark problems at the accuracy level $\epsilon = 1\mathrm{e}{-4}$}
\begin{tabular}{|l|l|l|l|l|l|l|l|l|}
\hline Function & \multicolumn{2}{|l|} { EODE } & \multicolumn{2}{|l|} { EODE-r }& \multicolumn{2}{l|} {EODE-b }&
\multicolumn{2}{l|} { EODE-rb } \\
 Index & PR & SR & PR & SR& PR & SR& PR & SR \\
 \hline
1 & \textbf{1}& 1&\textbf{1} & 1 & \textbf{1} & 1 &\textbf{1} & 1    \\
2 & \textbf{1}& 1&\textbf{1} & 1 & 0.88 & 0.4 &\textbf{1} & 1 \\
3 & \textbf{1}& 1&\textbf{1} & 1 & \textbf{1}& 1 &\textbf{1} & 1   \\
4 & \textbf{1}& 1&\textbf{1} & 1 & \textbf{1} & 1 &0.75 & 0.4  \\
5 &  \textbf{1}& 1&\textbf{1} & 1 & \textbf{1} & 1 &\textbf{1} & 1 \\
6 & 0.995 & 0.9 & \textbf{1}&\textbf{1}&	0.997&0&\textbf{1}&1\\
7 &  \textbf{0.805} & 0 &0.734&0&	0.733&0&0.788&0\\
8 & \textbf{0.845} & 0 & 0.567&0&	0.306&0&0.809&0\\
9 & \textbf{0.505} & 0 & 0.314&0&	0.452&0& 0.425&0\\
10 & \textbf{1} & 1 &\textbf{1}&1&	\textbf{1}&1&\textbf{1}&1\\
11 & \textbf{1} & 1 &\textbf{1}&1&	0.933&0.4& \textbf{1}&1\\
12 &0.975 & 0.8 & 0.925&0&	0.8&0&\textbf{1}&1\\
13 & \textbf{1} & 1 & 0.735&0&	0.55&0&0.667&0\\
14 & \textbf{0.8} & 0 & 0.667&0&	0.505&0&0.667&0\\
15 & \textbf{0.8} & 0 & 0.580&0&	0.612&0&0.785&0 \\
16 & \textbf{0.730} & 0&0.409&0&	0.441&0 &0.549&0\\
17 & \textbf{0.684} & 0 &0.186&0&	0.279&0&0.667&0 \\
18 &\textbf{0.684} & 0 & 0.147&0&	0&0&0.558&0\\
19 & \textbf{0.520} & 0 & 0&0&	0&0&0.395&0\\
20 & \textbf{0.250} & 0 & 0&0&	0&0&0.125&0\\
\hline bprs & \multicolumn{2}{|l|} {\textbf{18}} & \multicolumn{2}{l|} { 8}& \multicolumn{2}{l|} { 5 }& \multicolumn{2}{l|} { 8 }\\
\hline
\end{tabular}

\end{table}
The two different mutation operators are used with a probability of 0.5 each. The other components except mutation operators are not changed to note down the effects of mutation operators. The results corresponding to the usage of different mutation operators are shown in Table 8 at accuracy level $\epsilon$ = $1\mathrm{e}{-4}$.
 It is clear from Table 8 that EODE produces the best PR values for all benchmark functions except for the $6^{th}$ and $12^{th}$ functions. The performances of EODE-r and EODE-b are not so good as no peaks are found on the relatively high-dimensional functions.

Besides, Table 8 also shows that most of the compared algorithms perform well on functions 1-5, while EODE-b and EODE-rb are unable to find all the global peaks for functions 2 and 4 respectively. EODE-r finds all peaks in the functions 1-5. 
Here, we specifically compare the results of EODE and EODE-rb to illustrate the effect of three stage-wise mutation operators. As it can be referred from the previous experiment, the list functions are categorized in four different ways and the results on them are discussed below.\\
1) On the functions 1-5 which are considered to be simple, both EODE and EODE-rb locates all the optima except for function 4.\\
2) For functions 6-10 that consist of many global optima, it is clear that the performance of EODE is significantly better than that of EODE-rb except for function 6, which shows that the usage of mutation operators as per the stage of evolution process is highly effective. \\
3) For functions 11-15 which are low-dimensional composition functions, the PR values for EODE are superior to those of EODE-rb except for function 12. For functions 11-12, EODE-rb either performs better or equivalently well. For problems 13-15, EODE-rb does not perform well as compared to EODE.\\
4) On the relatively high-dimensional composition problems (the $16{th}–20^{th}$ problems), EODE performs better than EODE-rb, indicating that the mutation operations selected as per the stage of the evolution process can improve the overall performance of the algorithm.

\subsubsection{Different values of $\varphi_{1},\varphi_{2}$ }
In this section we have compared the results for different values of $\varphi_{1},\varphi_{2}$. In Section 3.1.1, $\varphi_{1},\varphi_{2}$ are used to form local sub-populations from global population. Thus, we set ($\varphi_{1},\varphi_{2}$) to three different values (1,1), (0.6,0.6), and (2,1) respectively. From Table 7 it can be seen that for $\varphi_{1}=1,\varphi_{2}=1$, EODE achieves the best overall results. For $\varphi_{1}=0.6,\varphi_{2}=0.6$, EODE achieves best result for function 15. For functions 1-5, $\varphi_{1}=2,\varphi_{2}=1$ achieves better results as compared to that of $\varphi_{1}=0.6,\varphi_{2}=0.6$. For functions 6-9 having large number of optima, $\varphi_{1}=0.6,\varphi_{2}=0.6$ achieves better results than $\varphi_{1}=2,\varphi_{2}=1$. However, for higher dimensional functions 16-20, EODE with $\varphi_{1}=2,\varphi_{2}=1$ seems to perform better except for functions 16 and 20. The behaviour is attributed to the number of species formed out of the global population. If the number of optima is large, then lower value of $\varphi_{1},\varphi_{2}$ is effective while if the number of optima is small, then higher value of $\varphi_{1},\varphi_{2}$ is effective.

\subsection{Different values of Jumping Rate (JR)}
Here, different values of JR are compared and it can be referred to from Table 9. In section 3.1.3, JR is used to form the opposite population. We modified the JR values to see if the opposite population in the exploration stage helps or they help in the exploitation stage.
\begin{table}
\renewcommand{\arraystretch}{1.2}
\centering
\caption{\label{tab:table-name}Experimental results of different values of $\varphi_{1}, \varphi_{2}$  on the benchmark problems at the accuracy level $\epsilon = 1\mathrm{e}{-4}$ }
\begin{tabular}{|l|l|l|l|l|l|l|}
\hline  Function&
\multicolumn{2}{|c|}{$\varphi_{1}$=1,$\varphi_{2}$=1} & \multicolumn{2}{|c|}{$\varphi_{1}$=0.6,$\varphi_{2}$=0.6} &\multicolumn{2}{|c|}{$\varphi_{1}$=2,$\varphi_{2}$=1 }\\

Index & PR & SR & PR & SR& PR & SR \\
\hline
1 & \textbf{1}& 1&\textbf{1} & 1 &\textbf{1} & 1    \\
2 & \textbf{1}& 1&0.68 & 0.2&0.92 & 0.6  \\
3 & \textbf{1}& 1&\textbf{1} & 1&\textbf{1} & 1    \\
4 & \textbf{1}& 1&0.5 & 1 &0.9 & 0.6 \\
5 &  \textbf{1}& 1&\textbf{1} & 1 &\textbf{1} & 1  \\
6 & 0.995 & 0.9 & \textbf{1} & 1& \textbf{1} & 1  \\
7 &  \textbf{0.805} & 0 &0.744&0& 0.688&0 \\
8 & \textbf{0.845} & 0 & 0.723&0&0.650&0\\
9 & \textbf{0.505} & 0 & 0.476&0&0.420&0\\
10 & \textbf{1} & 1 &\textbf{1}&1&\textbf{1}&1 \\
11 & \textbf{1} & 1 &\textbf{1} & 1&0.95&0.9\\
12 & \textbf{0.975} & 0.8& 0.95 & 0.4& \textbf{1} & 1\\
13 & \textbf{1} & 1 & 0.733&0&0.95&0.9\\
14 & \textbf{0.8} & 0 & 0.733&0&0.667&0 \\
15 & \textbf{0.8} & 0 & \textbf{1}&1&\textbf{0.8}&0 \\
16 & \textbf{0.730} & 0&0.709&0&0.667&0 \\
17 & \textbf{0.684} & 0 &0.515&0&0.580&0 \\
18 &\textbf{0.684} & 0 & 0.342&0&0.620&0\\
19 & \textbf{0.520} & 0 & 0.125&0 &0.418&0\\
20 & \textbf{0.250} & 0 & 0.175&0& 0.125&0\\
\hline bprs & \multicolumn{2}{|c|} {19} & \multicolumn{2}{c|} { 7 }& \multicolumn{2}{c|} { 7 }\\
\hline
\end{tabular}

\end{table}
JR is dependent upon the current generation and the maximum generation allowed using equation (12):
\begin{equation}
    JR=\frac{Gen}{MaxGen}
\end{equation}
$0\leq JR \leq 0.5$ indicates that the opposite population is generated in the early stage of evolution i.e. exploration while $0.67\leq JR \leq 1$ indicates that the opposite population is generated in the later stage of evolution i.e. exploitation. $0\leq JR \leq 1$ indicates that the opposite population is generated across the evolution stages. It can be seen that the opposite population generated in the later phase of evolution turns out to be helpful in locating more peaks. For the functions 1-5, EODE with $0.67\leq JR \leq 1$ and $0\leq JR \leq 1$ performs equivalently well. For functions 6-9, EODE with  $0\leq JR \leq 1$ performs better than EODE with  $0\leq JR \leq 0.5$ except the function 8. For functions 17 and 20, EODE with  $0\leq JR \leq 1$ achieves best results similar to that of EODE with  $0.67\leq JR \leq 1$.

\begin{table}
\renewcommand{\arraystretch}{1.2}
\centering
\caption{\label{tab:table-name}Experimental results of different values of JR  on the benchmark problems at the accuracy level $\epsilon = 1\mathrm{e}{-4}$ }
\begin{tabular}{|l|l|l|l|l|l|l|}
\hline Function & \multicolumn{2}{|l|}{0.67 $\leq$ JR $\leq$ 1} & \multicolumn{2}{|l|}{0 $\leq$ JR $\leq$ 0.5} &\multicolumn{2}{|l|}{0 $\leq$ JR $\leq$ 1 }\\
Index & PR & SR & PR & SR& PR & SR \\
\hline
1 & \textbf{1}& 1&\textbf{1} & 1 &\textbf{1} & 1    \\
2 & \textbf{1}& 1&0.84 & 0.6&\textbf{1}& 1  \\
3 & \textbf{1}& 1&\textbf{1} & 1&\textbf{1} & 1    \\
4 & \textbf{1}& 1&0.95 & 0.9 &\textbf{1} & 1  \\
5 &  \textbf{1}& 1&\textbf{1} & 1 &\textbf{1} & 1  \\
6 & 0.995 & 0.9 & \textbf{1} & 1& \textbf{1} & 1  \\
7 &  \textbf{0.805} & 0 &0.722&0& 0.768&0 \\
8 & \textbf{0.845} & 0 & 0.701&0&0.682&0\\
9 & \textbf{0.505} & 0 & 0.433&0&0.488&0\\
10 & \textbf{1} & 1 &\textbf{1}&1&0.9&0.4 \\
11 & \textbf{1} & 1 &\textbf{1} & 1&0.933&0.6\\
12 & \textbf{0.975} & 0.8& 0.95 & 0.4& 0.925&0.4\\
13 & \textbf{1} & 1 &0.866 & 0.2 &0.9&0.4\\
14 & \textbf{0.8} & 0 & 0.667&0&0.667&0 \\
15 & \textbf{0.8} & 0 & 0.667&0&0.6&0 \\
16 & \textbf{0.730} & 0&0.667&0&0.670&0 \\
17 & \textbf{0.684} & 0 &0.480&0&\textbf{0.684}&0 \\
18 &\textbf{0.684} & 0 & 0.486&0&0.470&0\\
19 & \textbf{0.520} & 0 & 0.225&0 &0.345&0\\
20 & \textbf{0.250} & 0 & 0.150&0& \textbf{0.250}&0\\
\hline bprs & \multicolumn{2}{|l|} {19} & \multicolumn{2}{l|} { 6 }& \multicolumn{2}{l|} { 8 }\\
\hline
\end{tabular}

\end{table}

\subsection{Convergence of population}
It is an important property of an evolutionary algorithm to converge the population towards optima.
The algorithm needs to maintain the found optima towards the end of evolution and still search for other optima. To assess the performance of EODE on converging the population towards optima, the solution distributions of some functions (i.e., F2, F4, F6, F7, F10, F11, F12, and F13), are presented on some specific generations. Figures 5-7 show the solution distribution of F2 with different generations (i.e., the generations are 1, 2, and 3). 
\clearpage
\begin{figure}[!htb]
\minipage{0.32\textwidth}
  \includegraphics[width=\linewidth]{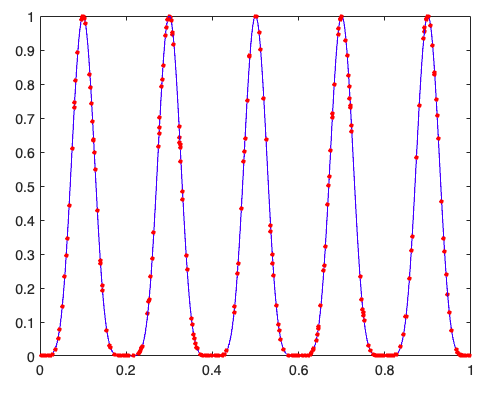}
  
  \caption{Gen-1 (F2)}\label{fig:awesome_image1}
\endminipage\hfill
\minipage{0.32\textwidth}
\includegraphics[width=\linewidth]{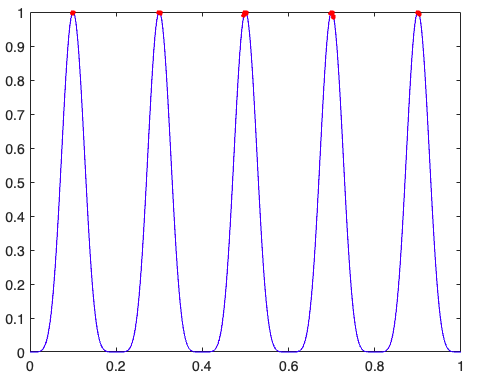}
  \caption{Gen-2 (F2)}\label{fig:awesome_image1}
\endminipage\hfill
\minipage{0.32\textwidth}%
 \includegraphics[width=\linewidth]{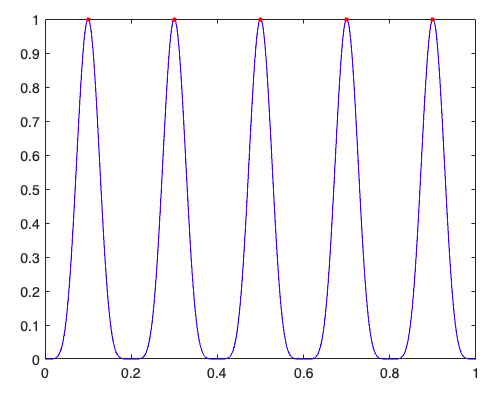}
  \caption{Gen-3 (F2)}\label{fig:awesome_image1}
\endminipage
\end{figure}
\begin{figure}[!htb]
\minipage{0.32\textwidth}
  \includegraphics[width=\linewidth]{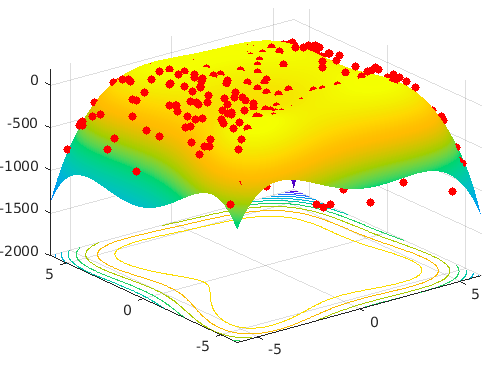}
  
  \caption{Gen-1 (F4)}\label{fig:awesome_image1}
\endminipage\hfill
\minipage{0.32\textwidth}
\includegraphics[width=\linewidth]{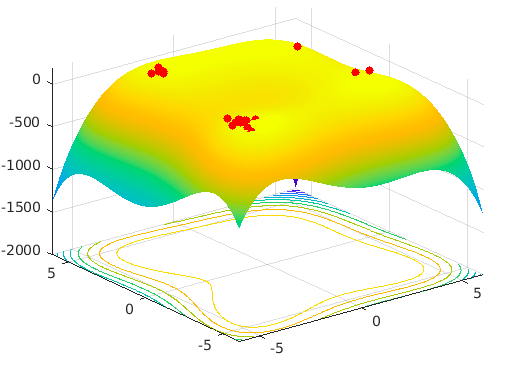}
  \caption{Gen-5 (F4)}\label{fig:awesome_image1}
\endminipage\hfill
\minipage{0.32\textwidth}%
 \includegraphics[width=\linewidth]{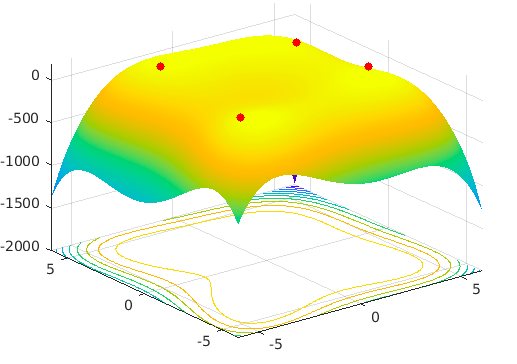}
  \caption{Gen-9 (F4)}\label{fig:awesome_image1}
\endminipage
\end{figure}

\begin{figure}[!htb]
\minipage{0.32\textwidth}
  \includegraphics[width=\linewidth]{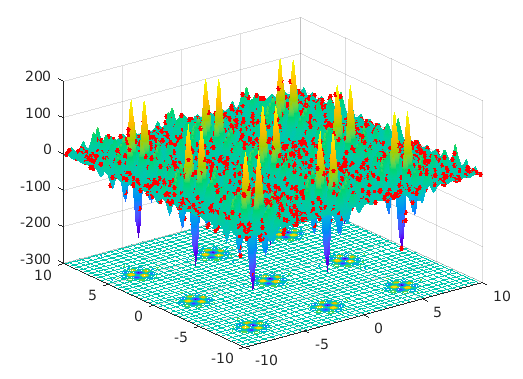}
  
  \caption{Gen-1 (F6)}\label{fig:awesome_image1}
\endminipage\hfill
\minipage{0.32\textwidth}
\includegraphics[width=\linewidth]{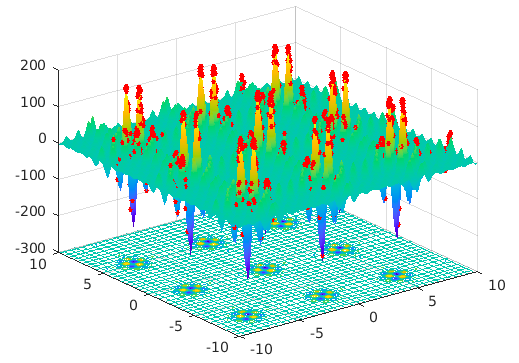}
  \caption{Gen-2 (F6)}\label{fig:awesome_image1}
\endminipage\hfill
\minipage{0.32\textwidth}%
 \includegraphics[width=\linewidth]{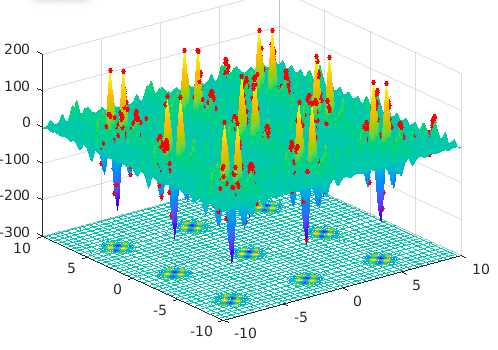}
  \caption{Gen-3 (F6)}\label{fig:awesome_image1}
\endminipage
\end{figure}
\begin{figure}[!htb]
\minipage{0.32\textwidth}
  \includegraphics[width=\linewidth]{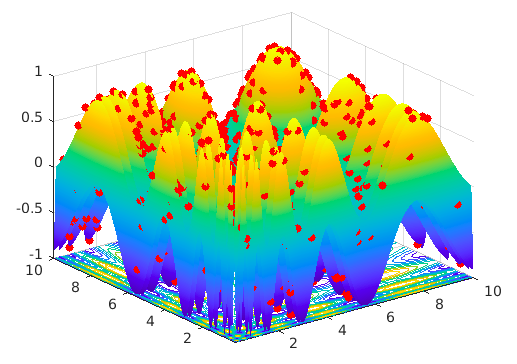}
  
  \caption{Gen-1 (F7)}\label{fig:awesome_image1}
\endminipage\hfill
\minipage{0.32\textwidth}
\includegraphics[width=\linewidth]{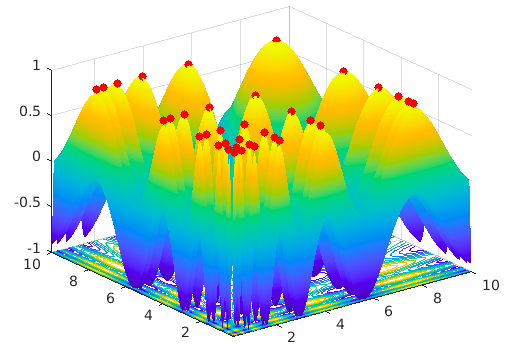}
  \caption{Gen-8 (F7)}\label{fig:awesome_image1}
\endminipage\hfill
\minipage{0.32\textwidth}%
 \includegraphics[width=\linewidth]{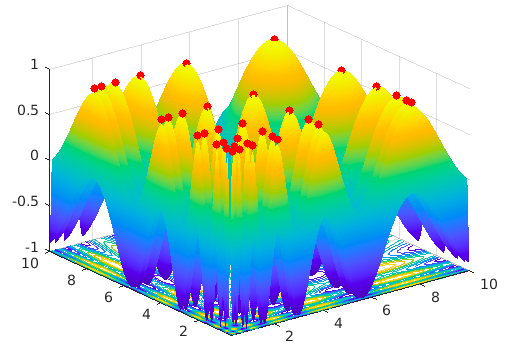}
  \caption{Gen-15 (F7)}\label{fig:awesome_image1}
\endminipage
\end{figure}
\begin{figure}[!htb]
\minipage{0.32\textwidth}
  \includegraphics[width=\linewidth]{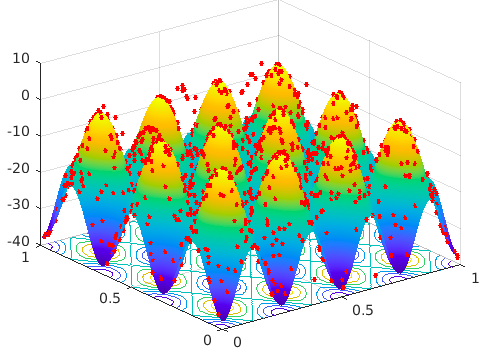}
  
  \caption{Gen-1 (F10)}\label{fig:awesome_image1}
\endminipage\hfill
\minipage{0.32\textwidth}
\includegraphics[width=\linewidth]{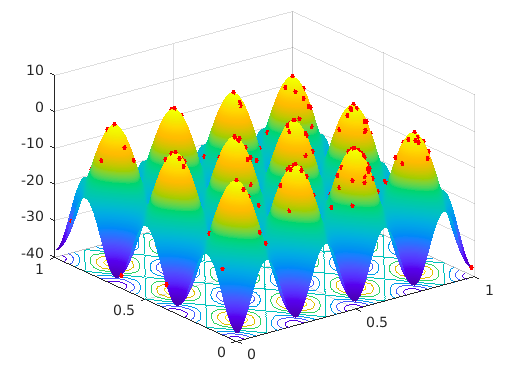}
  \caption{Gen-5 (F10)}\label{fig:awesome_image1}
\endminipage\hfill
\minipage{0.32\textwidth}%
 \includegraphics[width=\linewidth]{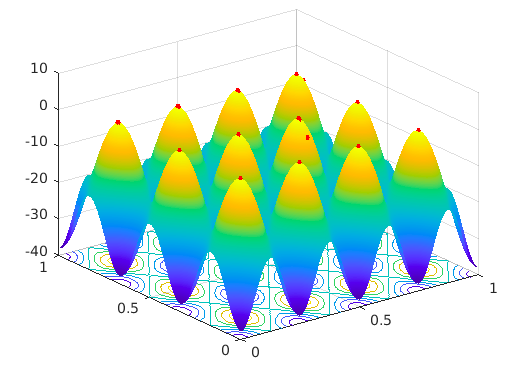}
  \caption{Gen-11 (F10)}\label{fig:awesome_image1}
\endminipage
\end{figure}
\begin{figure}[!htb]
\minipage{0.32\textwidth}
  \includegraphics[width=\linewidth]{figs/F11_1.png}
  
  \caption{Gen-1 (F11)}\label{fig:awesome_image1}
\endminipage\hfill
\minipage{0.32\textwidth}
\includegraphics[width=\linewidth]{figs/F11_2.png}
  \caption{Gen-8 (F11)}\label{fig:awesome_image1}
\endminipage\hfill
\minipage{0.32\textwidth}%
 \includegraphics[width=\linewidth]{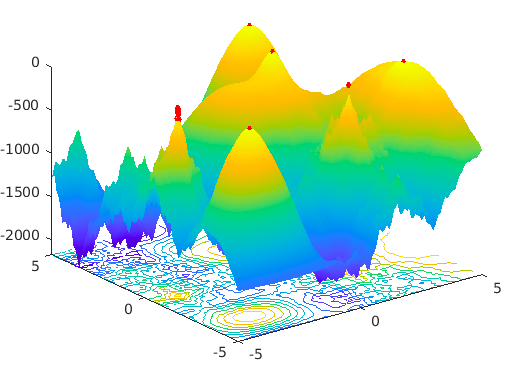}
  \caption{Gen-14 (F11)}\label{fig:awesome_image1}
\endminipage
\end{figure}
\begin{figure}[!htb]
\minipage{0.32\textwidth}
  \includegraphics[width=\linewidth]{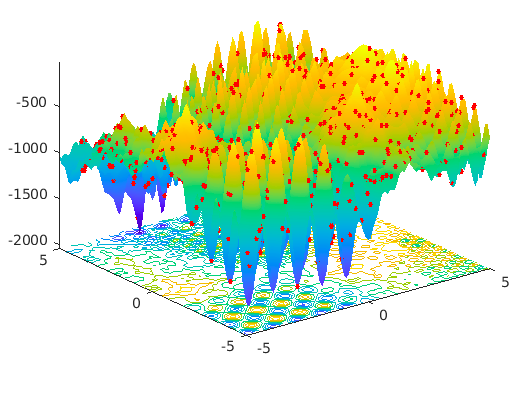}
  
  \caption{Gen-1 (F12)}\label{fig:awesome_image1}
\endminipage\hfill
\minipage{0.32\textwidth}
\includegraphics[width=\linewidth]{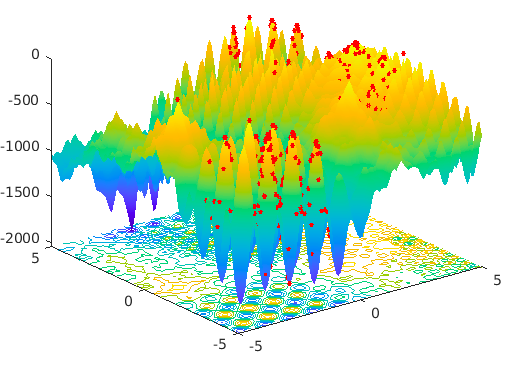}
  \caption{Gen-5 (F12)}\label{fig:awesome_image1}
\endminipage\hfill
\minipage{0.32\textwidth}%
 \includegraphics[width=\linewidth]{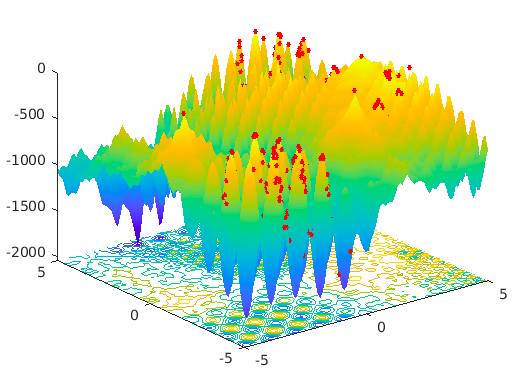}
  \caption{Gen-11 (F12)}\label{fig:awesome_image1}
\endminipage
\end{figure}
\begin{figure}[!htb]
\minipage{0.32\textwidth}
  \includegraphics[width=\linewidth]{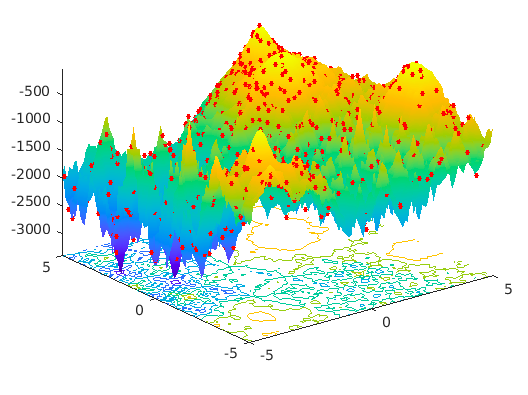}
  
  \caption{Gen-1 (F13)}\label{fig:awesome_image1}
\endminipage\hfill
\minipage{0.32\textwidth}
\includegraphics[width=\linewidth]{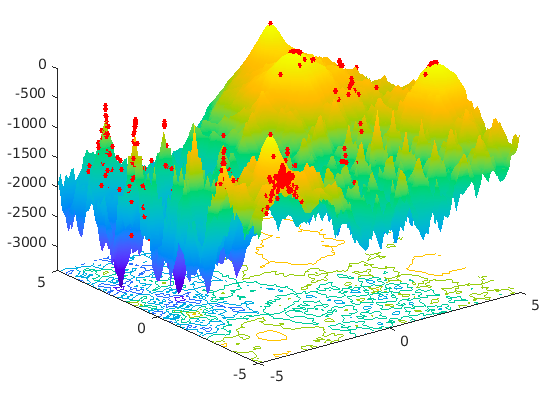}
  \caption{Gen-5 (F13)}\label{fig:awesome_image1}
\endminipage\hfill
\minipage{0.32\textwidth}%
 \includegraphics[width=\linewidth]{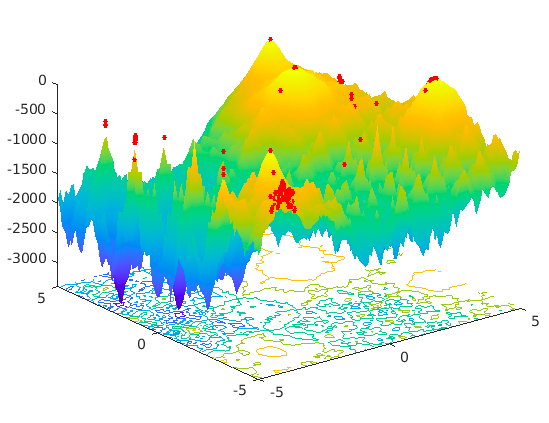}
  \caption{Gen-11 (F13)}\label{fig:awesome_image1}
\endminipage
\end{figure}
It can be clearly seen that EODE
has located all of the global optima when the generation is 3 on F2 [i.e., Figure 7]. Hence, it can be concluded that EODE can locate the global optima quickly which indicates that EODE has a good ability of global search. It is important to note that the generation here refers to the number of times a multi-species framework is performed given in Algorithm 2. It does not refer to the generation of DE operations directly. Figures 8-10 show the solution distribution of F4 with different generations (i.e., 1, 5, and 9). It can be inferred from Figure 10 that the population converges completely in the $9^{th}$ generation. Similarly, for F7, F10, EODE obtained complete convergence. However, for F6 (figs 11-13), F11( figs 20-22), F12(figs 23-25), and F13 (figs 26-28) there is some dispersion from the peak but still, the convergence is nearly complete.  Hence, it can be concluded that as the generation increases the solutions move towards a convergence state.
Also, it is clear that the population does not diverge from the optima with the increasing evolution in EODE. Hence, our EODE can converge towards the global optima by the end of evolution.

\section{Conclusion}
In this paper, we have proposed EODE to handle MMOPs, where several species techniques have been adopted to simultaneously find multiple optima. In EODE, two-level NBC-minsize has been used to divide the population into several species to locate the optima. To eliminate the drawbacks of NBC-minsize an additional check is also performed to improve its efficacy. Besides, a species balance strategy has been proposed to balance the species having uneven sizes. The modified opposition differential evolution algorithm acts as the evolutionary process. The evolutionary process is further divided virtually into three stages and mutation operations for each of the stages are embedded individually. Besides, an adaptive parameter strategy has also been proposed to learn the parameters of the algorithm dynamically. We have also proposed a local search method to improve the quality of candidate solutions. Finally, EODE has been compared with several state-of-the-art algorithms. The experimental results show that EODE performed better than other algorithms on most multimodal benchmark problems.
However, in our experiments, we have achieved better results for problems having a moderate number of global optima, but it still needs to improve on highly multimodal functions. Hence, we aim to improve EODE in our future work and test it on highly multimodal problems.

\section{Acknowledgement}
The authors would like to thank Xin Lin and Wenjian Luo for providing source code of \cite{NBC-Minsize} that provided some helpful insights into the design and working of the proposed algorithm.

\section{Declarations}
\subsection{Funding and/or Conflicts of interests/Competing interests/Data Availability Statements}
This work is part of an academic course and it has not been funded by any organization. The author(s) declare no competing interests. The dataset can be found at \url{https://github.com/mikeagn/CEC2013}.

\bibliography{sn-bibliography}


\begin{thebibliography}{39}
\ifx \bisbn   \undefined \def \bisbn  #1{ISBN #1}\fi
\ifx \binits  \undefined \def \binits#1{#1}\fi
\ifx \bauthor  \undefined \def \bauthor#1{#1}\fi
\ifx \batitle  \undefined \def \batitle#1{#1}\fi
\ifx \bjtitle  \undefined \def \bjtitle#1{#1}\fi
\ifx \bvolume  \undefined \def \bvolume#1{\textbf{#1}}\fi
\ifx \byear  \undefined \def \byear#1{#1}\fi
\ifx \bissue  \undefined \def \bissue#1{#1}\fi
\ifx \bfpage  \undefined \def \bfpage#1{#1}\fi
\ifx \blpage  \undefined \def \blpage #1{#1}\fi
\ifx \burl  \undefined \def \burl#1{\textsf{#1}}\fi
\ifx \doiurl  \undefined \def \doiurl#1{\url{https://doi.org/#1}}\fi
\ifx \betal  \undefined \def \betal{\textit{et al.}}\fi
\ifx \binstitute  \undefined \def \binstitute#1{#1}\fi
\ifx \binstitutionaled  \undefined \def \binstitutionaled#1{#1}\fi
\ifx \bctitle  \undefined \def \bctitle#1{#1}\fi
\ifx \beditor  \undefined \def \beditor#1{#1}\fi
\ifx \bpublisher  \undefined \def \bpublisher#1{#1}\fi
\ifx \bbtitle  \undefined \def \bbtitle#1{#1}\fi
\ifx \bedition  \undefined \def \bedition#1{#1}\fi
\ifx \bseriesno  \undefined \def \bseriesno#1{#1}\fi
\ifx \blocation  \undefined \def \blocation#1{#1}\fi
\ifx \bsertitle  \undefined \def \bsertitle#1{#1}\fi
\ifx \bsnm \undefined \def \bsnm#1{#1}\fi
\ifx \bsuffix \undefined \def \bsuffix#1{#1}\fi
\ifx \bparticle \undefined \def \bparticle#1{#1}\fi
\ifx \barticle \undefined \def \barticle#1{#1}\fi
\bibcommenthead
\ifx \bconfdate \undefined \def \bconfdate #1{#1}\fi
\ifx \botherref \undefined \def \botherref #1{#1}\fi
\ifx \url \undefined \def \url#1{\textsf{#1}}\fi
\ifx \bchapter \undefined \def \bchapter#1{#1}\fi
\ifx \bbook \undefined \def \bbook#1{#1}\fi
\ifx \bcomment \undefined \def \bcomment#1{#1}\fi
\ifx \oauthor \undefined \def \oauthor#1{#1}\fi
\ifx \citeauthoryear \undefined \def \citeauthoryear#1{#1}\fi
\ifx \endbibitem  \undefined \def \endbibitem {}\fi
\ifx \bconflocation  \undefined \def \bconflocation#1{#1}\fi
\ifx \arxivurl  \undefined \def \arxivurl#1{\textsf{#1}}\fi
\csname PreBibitemsHook\endcsname

\bibitem{app1}
\begin{bchapter}
\bauthor{\bsnm{Qing}, \binits{L.}},
\bauthor{\bsnm{Gang}, \binits{W.}},
\bauthor{\bsnm{Qiuping}, \binits{W.}}:
\bctitle{Restricted evolution based multimodal function optimization in
  holographic grating design}.
In: \bbtitle{2005 IEEE Congress on Evolutionary Computation},
vol. \bseriesno{1},
pp. \bfpage{789}--\blpage{7941}
(\byear{2005}).
\doiurl{10.1109/CEC.2005.1554763}
\end{bchapter}
\endbibitem

\bibitem{app2}
\begin{barticle}
\bauthor{\bsnm{Woo}, \binits{D.-K.}},
\bauthor{\bsnm{Choi}, \binits{J.-H.}},
\bauthor{\bsnm{Ali}, \binits{M.}},
\bauthor{\bsnm{Jung}, \binits{H.-K.}}:
\batitle{A novel multimodal optimization algorithm applied to electromagnetic
  optimization}.
\bjtitle{IEEE Transactions on Magnetics}
\bvolume{47}(\bissue{6}),
\bfpage{1667}--\blpage{1673}
(\byear{2011}).
\doiurl{10.1109/TMAG.2011.2106218}
\end{barticle}
\endbibitem

\bibitem{app3}
\begin{bchapter}
\bauthor{\bsnm{Wong}, \binits{K.-C.}},
\bauthor{\bsnm{Leung}, \binits{K.}},
\bauthor{\bsnm{Wong}, \binits{M.H.}}:
\bctitle{Protein structure prediction on a lattice model via multimodal
  optimization techniques},
vol. \bseriesno{2010},
pp. \bfpage{155}--\blpage{162}
(\byear{2010}).
\doiurl{10.1145/1830483.1830513}
\end{bchapter}
\endbibitem

\bibitem{app4}
\begin{bchapter}
\bauthor{\bsnm{Boughanem}, \binits{M.}},
\bauthor{\bsnm{Tamine}, \binits{L.}}:
\bctitle{A study on using genetic niching for query optimisation in document
  retrieval}.
In: \beditor{\bsnm{Crestani}, \binits{F.}},
\beditor{\bsnm{Girolami}, \binits{M.}},
\beditor{\bparticle{van} \bsnm{Rijsbergen}, \binits{C.J.}} (eds.)
\bbtitle{Advances in Information Retrieval},
pp. \bfpage{135}--\blpage{149}.
\bpublisher{Springer},
\blocation{Berlin, Heidelberg}
(\byear{2002}).
\doiurl{10.1007/3-540-45886-7_10}
\end{bchapter}
\endbibitem

\bibitem{EAs}
\begin{barticle}
\bauthor{\bsnm{Covantes~Osuna}, \binits{E.}},
\bauthor{\bsnm{Sudholt}, \binits{D.}}:
\batitle{Runtime analysis of crowding mechanisms for multimodal optimization}.
\bjtitle{IEEE Transactions on Evolutionary Computation}
\bvolume{24}(\bissue{3}),
\bfpage{581}--\blpage{592}
(\byear{2020}).
\doiurl{10.1109/TEVC.2019.2914606}
\end{barticle}
\endbibitem

\bibitem{Swarm}
\begin{barticle}
\bauthor{\bsnm{Li}, \binits{Y.}},
\bauthor{\bsnm{Zhan}, \binits{Z.-H.}},
\bauthor{\bsnm{Lin}, \binits{S.}},
\bauthor{\bsnm{Zhang}, \binits{J.}},
\bauthor{\bsnm{Luo}, \binits{X.}}:
\batitle{Competitive and cooperative particle swarm optimization with
  information sharing mechanism for global optimization problems}.
\bjtitle{Information Sciences}
\bvolume{293},
\bfpage{370}--\blpage{382}
(\byear{2015}).
\doiurl{10.1016/j.ins.2014.09.030}
\end{barticle}
\endbibitem

\bibitem{GA}
\begin{bchapter}
\bauthor{\bsnm{Yao}, \binits{J.}},
\bauthor{\bsnm{Kharma}, \binits{N.}},
\bauthor{\bsnm{Grogono}, \binits{P.}}:
\bctitle{Bmpga: a bi-objective multi-population genetic algorithm for
  multi-modal function optimization}.
In: \bbtitle{2005 IEEE Congress on Evolutionary Computation},
vol. \bseriesno{1},
pp. \bfpage{816}--\blpage{8231}
(\byear{2005}).
\doiurl{10.1109/CEC.2005.1554767}
\end{bchapter}
\endbibitem

\bibitem{ACO}
\begin{barticle}
\bauthor{\bsnm{Yang}, \binits{Q.}},
\bauthor{\bsnm{Chen}, \binits{W.-N.}},
\bauthor{\bsnm{Yu}, \binits{Z.}},
\bauthor{\bsnm{Gu}, \binits{T.}},
\bauthor{\bsnm{Li}, \binits{Y.}},
\bauthor{\bsnm{Zhang}, \binits{H.}},
\bauthor{\bsnm{Zhang}, \binits{J.}}:
\batitle{Adaptive multimodal continuous ant colony optimization}.
\bjtitle{IEEE Transactions on Evolutionary Computation}
\bvolume{21}(\bissue{2}),
\bfpage{191}--\blpage{205}
(\byear{2017}).
\doiurl{10.1109/TEVC.2016.2591064}
\end{barticle}
\endbibitem

\bibitem{EDA}
\begin{barticle}
\bauthor{\bsnm{Li}, \binits{Y.}},
\bauthor{\bsnm{Zhan}, \binits{Z.-H.}},
\bauthor{\bsnm{Lin}, \binits{S.}},
\bauthor{\bsnm{Zhang}, \binits{J.}},
\bauthor{\bsnm{Luo}, \binits{X.}}:
\batitle{Competitive and cooperative particle swarm optimization with
  information sharing mechanism for global optimization problems}.
\bjtitle{Information Sciences}
\bvolume{293},
\bfpage{370}--\blpage{382}
(\byear{2015}).
\doiurl{10.1016/j.ins.2014.09.030}
\end{barticle}
\endbibitem

\bibitem{PSO}
\begin{barticle}
\bauthor{\bsnm{Cao}, \binits{Y.}},
\bauthor{\bsnm{Zhang}, \binits{H.}},
\bauthor{\bsnm{Li}, \binits{W.}},
\bauthor{\bsnm{Zhou}, \binits{M.}},
\bauthor{\bsnm{Zhang}, \binits{Y.}},
\bauthor{\bsnm{Chaovalitwongse}, \binits{W.A.}}:
\batitle{Comprehensive learning particle swarm optimization algorithm with
  local search for multimodal functions}.
\bjtitle{IEEE Transactions on Evolutionary Computation}
\bvolume{23}(\bissue{4}),
\bfpage{718}--\blpage{731}
(\byear{2019}).
\doiurl{10.1109/TEVC.2018.2885075}
\end{barticle}
\endbibitem

\bibitem{DE}
\begin{barticle}
\bauthor{\bsnm{Wang}, \binits{Z.-J.}},
\bauthor{\bsnm{Zhan}, \binits{Z.-H.}},
\bauthor{\bsnm{Lin}, \binits{Y.}},
\bauthor{\bsnm{Yu}, \binits{W.-J.}},
\bauthor{\bsnm{Wang}, \binits{H.}},
\bauthor{\bsnm{Kwong}, \binits{S.}},
\bauthor{\bsnm{Zhang}, \binits{J.}}:
\batitle{Automatic niching differential evolution with contour prediction
  approach for multimodal optimization problems}.
\bjtitle{IEEE Transactions on Evolutionary Computation}
\bvolume{24}(\bissue{1}),
\bfpage{114}--\blpage{128}
(\byear{2020}).
\doiurl{10.1109/TEVC.2019.2910721}
\end{barticle}
\endbibitem

\bibitem{niching}
\begin{botherref}
\oauthor{\bsnm{Mahfoud}, \binits{S.W.}}:
Niching methods for genetic algorithms.
PhD thesis,
USA
(1996).
UMI Order No. GAX95-43663
\end{botherref}
\endbibitem

\bibitem{niching1}
\begin{botherref}
De jong, k. a. (1975). an analysis of the behavior of a class of genetic
  adaptive systems (doc-.
(2007)
\end{botherref}
\endbibitem

\bibitem{niching2}
\begin{bchapter}
\bauthor{\bsnm{Petrowski}, \binits{A.}}:
\bctitle{A clearing procedure as a niching method for genetic algorithms}.
In: \bbtitle{Proceedings of IEEE International Conference on Evolutionary
  Computation},
pp. \bfpage{798}--\blpage{803}
(\byear{1996}).
\doiurl{10.1109/ICEC.1996.542703}
\end{bchapter}
\endbibitem

\bibitem{niching3}
\begin{bbook}
\bauthor{\bsnm{Holland}, \binits{J.H.}}:
\bbtitle{{Adaptation in Natural and Artificial Systems: An Introductory
  Analysis with Applications to Biology, Control, and Artificial
  Intelligence}}.
\bpublisher{The MIT Press}, \blocation{???}
(\byear{1992}).
\doiurl{10.7551/mitpress/1090.001.0001}.
\burl{https://doi.org/10.7551/mitpress/1090.001.0001}
\end{bbook}
\endbibitem

\bibitem{niching4}
\begin{barticle}
\bauthor{\bsnm{Della~Cioppa}, \binits{A.}},
\bauthor{\bsnm{De~Stefano}, \binits{C.}},
\bauthor{\bsnm{Marcelli}, \binits{A.}}:
\batitle{Where are the niches? dynamic fitness sharing}.
\bjtitle{IEEE Transactions on Evolutionary Computation}
\bvolume{11}(\bissue{4}),
\bfpage{453}--\blpage{465}
(\byear{2007}).
\doiurl{10.1109/TEVC.2006.882433}
\end{barticle}
\endbibitem

\bibitem{topological}
\begin{barticle}
\bauthor{\bsnm{Stoean}, \binits{C.}},
\bauthor{\bsnm{Preuss}, \binits{M.}},
\bauthor{\bsnm{Stoean}, \binits{R.}},
\bauthor{\bsnm{Dumitrescu}, \binits{D.}}:
\batitle{Multimodal optimization by means of a topological species conservation
  algorithm}.
\bjtitle{IEEE Transactions on Evolutionary Computation}
\bvolume{14}(\bissue{6}),
\bfpage{842}--\blpage{864}
(\byear{2010}).
\doiurl{10.1109/TEVC.2010.2041668}
\end{barticle}
\endbibitem

\bibitem{clustering}
\begin{barticle}
\bauthor{\bsnm{Gao}, \binits{W.}},
\bauthor{\bsnm{Yen}, \binits{G.G.}},
\bauthor{\bsnm{Liu}, \binits{S.}}:
\batitle{A cluster-based differential evolution with self-adaptive strategy for
  multimodal optimization}.
\bjtitle{IEEE Transactions on Cybernetics}
\bvolume{44}(\bissue{8}),
\bfpage{1314}--\blpage{1327}
(\byear{2014}).
\doiurl{10.1109/TCYB.2013.2282491}
\end{barticle}
\endbibitem

\bibitem{ada1}
\begin{barticle}
\bauthor{\bsnm{Yao}, \binits{J.}},
\bauthor{\bsnm{Kharma}, \binits{N.}},
\bauthor{\bsnm{Grogono}, \binits{P.}}:
\batitle{Bi-objective multipopulation genetic algorithm for multimodal function
  optimization}.
\bjtitle{IEEE Transactions on Evolutionary Computation}
\bvolume{14}(\bissue{1}),
\bfpage{80}--\blpage{102}
(\byear{2010}).
\doiurl{10.1109/TEVC.2009.2017517}
\end{barticle}
\endbibitem

\bibitem{ada2}
\begin{barticle}
\bauthor{\bsnm{Biswas}, \binits{S.}},
\bauthor{\bsnm{Kundu}, \binits{S.}},
\bauthor{\bsnm{Das}, \binits{S.}}:
\batitle{Inducing niching behavior in differential evolution through local
  information sharing}.
\bjtitle{IEEE Transactions on Evolutionary Computation}
\bvolume{19}(\bissue{2}),
\bfpage{246}--\blpage{263}
(\byear{2015}).
\doiurl{10.1109/TEVC.2014.2313659}
\end{barticle}
\endbibitem

\bibitem{ada3}
\begin{barticle}
\bauthor{\bsnm{Li}, \binits{X.}}:
\batitle{Niching without niching parameters: Particle swarm optimization using
  a ring topology}.
\bjtitle{IEEE Transactions on Evolutionary Computation}
\bvolume{14}(\bissue{1}),
\bfpage{150}--\blpage{169}
(\byear{2010}).
\doiurl{10.1109/TEVC.2009.2026270}
\end{barticle}
\endbibitem

\bibitem{b1}
\begin{barticle}
\bauthor{\bsnm{Rahnamayan}, \binits{S.}},
\bauthor{\bsnm{Tizhoosh}, \binits{H.R.}},
\bauthor{\bsnm{Salama}, \binits{M.M.A.}}:
\batitle{Opposition-based differential evolution}.
\bjtitle{IEEE Transactions on Evolutionary Computation}
\bvolume{12}(\bissue{1}),
\bfpage{64}--\blpage{79}
(\byear{2008}).
\doiurl{10.1109/TEVC.2007.894200}
\end{barticle}
\endbibitem

\bibitem{NBC}
\begin{bchapter}
\bauthor{\bsnm{Preuss}, \binits{M.}}:
\bctitle{Niching the cma-es via nearest-better clustering},
pp. \bfpage{1711}--\blpage{1718}
(\byear{2010}).
\doiurl{10.1145/1830761.1830793}
\end{bchapter}
\endbibitem

\bibitem{NBC-Minsize}
\begin{barticle}
\bauthor{\bsnm{Lin}, \binits{X.}},
\bauthor{\bsnm{Luo}, \binits{W.}},
\bauthor{\bsnm{Xu}, \binits{P.}}:
\batitle{Differential evolution for multimodal optimization with species by
  nearest-better clustering}.
\bjtitle{IEEE Transactions on Cybernetics}
\bvolume{51}(\bissue{2}),
\bfpage{970}--\blpage{983}
(\byear{2021}).
\doiurl{10.1109/TCYB.2019.2907657}
\end{barticle}
\endbibitem

\bibitem{multi-pop}
\begin{botherref}
\oauthor{\bsnm{Zaharie}, \binits{D.}}:
A multipopulation differential evolution algorithm for multimodal optimization
(2004)
\end{botherref}
\endbibitem

\bibitem{FBK-DE}
\begin{barticle}
\bauthor{\bsnm{Lin}, \binits{X.}},
\bauthor{\bsnm{Luo}, \binits{W.}},
\bauthor{\bsnm{Xu}, \binits{P.}}:
\batitle{Differential evolution for multimodal optimization with species by
  nearest-better clustering}.
\bjtitle{IEEE Transactions on Cybernetics}
\bvolume{51}(\bissue{2}),
\bfpage{970}--\blpage{983}
(\byear{2021}).
\doiurl{10.1109/TCYB.2019.2907657}
\end{barticle}
\endbibitem

\bibitem{SHADE}
\begin{bchapter}
\bauthor{\bsnm{Tanabe}, \binits{R.}},
\bauthor{\bsnm{Fukunaga}, \binits{A.}}:
\bctitle{Success-history based parameter adaptation for differential
  evolution}.
In: \bbtitle{2013 IEEE Congress on Evolutionary Computation},
pp. \bfpage{71}--\blpage{78}
(\byear{2013}).
\doiurl{10.1109/CEC.2013.6557555}
\end{bchapter}
\endbibitem

\bibitem{Wangg}
\begin{barticle}
\bauthor{\bsnm{Wang}, \binits{Z.-J.}},
\bauthor{\bsnm{Zhan}, \binits{Z.-H.}},
\bauthor{\bsnm{Lin}, \binits{Y.}},
\bauthor{\bsnm{Yu}, \binits{W.-J.}},
\bauthor{\bsnm{Yuan}, \binits{H.-Q.}},
\bauthor{\bsnm{Gu}, \binits{T.-L.}},
\bauthor{\bsnm{Kwong}, \binits{S.}},
\bauthor{\bsnm{Zhang}, \binits{J.}}:
\batitle{Dual-strategy differential evolution with affinity propagation
  clustering for multimodal optimization problems}.
\bjtitle{IEEE Transactions on Evolutionary Computation}
\bvolume{22}(\bissue{6}),
\bfpage{894}--\blpage{908}
(\byear{2018}).
\doiurl{10.1109/TEVC.2017.2769108}
\end{barticle}
\endbibitem

\bibitem{Yangg}
\begin{barticle}
\bauthor{\bsnm{Yang}, \binits{Q.}},
\bauthor{\bsnm{Chen}, \binits{W.-N.}},
\bauthor{\bsnm{Yu}, \binits{Z.}},
\bauthor{\bsnm{Gu}, \binits{T.}},
\bauthor{\bsnm{Li}, \binits{Y.}},
\bauthor{\bsnm{Zhang}, \binits{H.}},
\bauthor{\bsnm{Zhang}, \binits{J.}}:
\batitle{Adaptive multimodal continuous ant colony optimization}.
\bjtitle{IEEE Transactions on Evolutionary Computation}
\bvolume{21}(\bissue{2}),
\bfpage{191}--\blpage{205}
(\byear{2017}).
\doiurl{10.1109/TEVC.2016.2591064}
\end{barticle}
\endbibitem

\bibitem{crowding}
\begin{bchapter}
\bauthor{\bsnm{Thomsen}, \binits{R.}}:
\bctitle{Multimodal optimization using crowding-based differential evolution}.
In: \bbtitle{Proceedings of the 2004 Congress on Evolutionary Computation (IEEE
  Cat. No.04TH8753)},
vol. \bseriesno{2},
pp. \bfpage{1382}--\blpage{13892}
(\byear{2004}).
\doiurl{10.1109/CEC.2004.1331058}
\end{bchapter}
\endbibitem

\bibitem{DE5}
\begin{bchapter}
\bauthor{\bsnm{Li}, \binits{X.}}:
\bctitle{Efficient differential evolution using speciation for multimodal
  function optimization},
pp. \bfpage{873}--\blpage{880}
(\byear{2005}).
\doiurl{10.1145/1068009.1068156}
\end{bchapter}
\endbibitem

\bibitem{Qu-DE}
\begin{barticle}
\bauthor{\bsnm{Qu}, \binits{B.Y.}},
\bauthor{\bsnm{Suganthan}, \binits{P.N.}},
\bauthor{\bsnm{Liang}, \binits{J.J.}}:
\batitle{Differential evolution with neighborhood mutation for multimodal
  optimization}.
\bjtitle{IEEE Transactions on Evolutionary Computation}
\bvolume{16}(\bissue{5}),
\bfpage{601}--\blpage{614}
(\byear{2012}).
\doiurl{10.1109/TEVC.2011.2161873}
\end{barticle}
\endbibitem

\bibitem{MOMMOP}
\begin{barticle}
\bauthor{\bsnm{Wang}, \binits{Y.}},
\bauthor{\bsnm{Li}, \binits{H.-X.}},
\bauthor{\bsnm{Yen}, \binits{G.G.}},
\bauthor{\bsnm{Song}, \binits{W.}}:
\batitle{Mommop: Multiobjective optimization for locating multiple optimal
  solutions of multimodal optimization problems}.
\bjtitle{IEEE Transactions on Cybernetics}
\bvolume{45}(\bissue{4}),
\bfpage{830}--\blpage{843}
(\byear{2015}).
\doiurl{10.1109/TCYB.2014.2337117}
\end{barticle}
\endbibitem

\bibitem{pnpcde}
\begin{barticle}
\bauthor{\bsnm{Biswas}, \binits{S.}},
\bauthor{\bsnm{Kundu}, \binits{S.}},
\bauthor{\bsnm{Das}, \binits{S.}}:
\batitle{An improved parent-centric mutation with normalized neighborhoods for
  inducing niching behavior in differential evolution}.
\bjtitle{IEEE Transactions on Cybernetics}
\bvolume{44}(\bissue{10}),
\bfpage{1726}--\blpage{1737}
(\byear{2014}).
\doiurl{10.1109/TCYB.2013.2292971}
\end{barticle}
\endbibitem

\bibitem{Qu}
\begin{barticle}
\bauthor{\bsnm{Qu}, \binits{B.Y.}},
\bauthor{\bsnm{Suganthan}, \binits{P.N.}},
\bauthor{\bsnm{Das}, \binits{S.}}:
\batitle{A distance-based locally informed particle swarm model for multimodal
  optimization}.
\bjtitle{IEEE Transactions on Evolutionary Computation}
\bvolume{17}(\bissue{3}),
\bfpage{387}--\blpage{402}
(\byear{2013}).
\doiurl{10.1109/TEVC.2012.2203138}
\end{barticle}
\endbibitem

\bibitem{DEcl}
\begin{bchapter}
\bauthor{\bsnm{Bošković}, \binits{B.}},
\bauthor{\bsnm{Brest}, \binits{J.}}:
\bctitle{Clustering and differential evolution for multimodal optimization}.
In: \bbtitle{2017 IEEE Congress on Evolutionary Computation (CEC)},
pp. \bfpage{698}--\blpage{705}
(\byear{2017}).
\doiurl{10.1109/CEC.2017.7969378}
\end{bchapter}
\endbibitem

\bibitem{Zhao}
\begin{barticle}
\bauthor{\bsnm{Zhao}, \binits{H.}},
\bauthor{\bsnm{Zhan}, \binits{Z.-H.}},
\bauthor{\bsnm{Lin}, \binits{Y.}},
\bauthor{\bsnm{Chen}, \binits{X.}},
\bauthor{\bsnm{Luo}, \binits{X.-N.}},
\bauthor{\bsnm{Zhang}, \binits{J.}},
\bauthor{\bsnm{Kwong}, \binits{S.}},
\bauthor{\bsnm{Zhang}, \binits{J.}}:
\batitle{Local binary pattern-based adaptive differential evolution for
  multimodal optimization problems}.
\bjtitle{IEEE Transactions on Cybernetics}
\bvolume{50}(\bissue{7}),
\bfpage{3343}--\blpage{3357}
(\byear{2020}).
\doiurl{10.1109/TCYB.2019.2927780}
\end{barticle}
\endbibitem

\bibitem{MaHDE}
\begin{barticle}
\bauthor{\bsnm{Hong}, \binits{Z.}},
\bauthor{\bsnm{Chen}, \binits{Z.-G.}},
\bauthor{\bsnm{Liu}, \binits{D.}},
\bauthor{\bsnm{Zhan}, \binits{Z.-H.}},
\bauthor{\bsnm{Zhang}, \binits{J.}}:
\batitle{A multi-angle hierarchical differential evolution approach for
  multimodal optimization problems}.
\bjtitle{IEEE Access}
\bvolume{8},
\bfpage{178322}--\blpage{178335}
(\byear{2020}).
\doiurl{10.1109/ACCESS.2020.3027559}
\end{barticle}
\endbibitem

\bibitem{mutation}
\begin{barticle}
\bauthor{\bsnm{Islam}, \binits{S.M.}},
\bauthor{\bsnm{Das}, \binits{S.}},
\bauthor{\bsnm{Ghosh}, \binits{S.}},
\bauthor{\bsnm{Roy}, \binits{S.}},
\bauthor{\bsnm{Suganthan}, \binits{P.N.}}:
\batitle{An adaptive differential evolution algorithm with novel mutation and
  crossover strategies for global numerical optimization}.
\bjtitle{IEEE Transactions on Systems, Man, and Cybernetics, Part B
  (Cybernetics)}
\bvolume{42}(\bissue{2}),
\bfpage{482}--\blpage{500}
(\byear{2012}).
\doiurl{10.1109/TSMCB.2011.2167966}
\end{barticle}
\endbibitem

\end{thebibliography}


\end{document}